\def\uselab{}

\documentclass{article} % For LaTeX2e
\ifdefined\uselab
  \usepackage{labtemplate/aditi}
  \usepackage{labtemplate/utils}
  \usepackage{labtemplate/algorithm}
  \usepackage{labtemplate/algorithmic}
\else
  \usepackage{iclr2027_conference,times}
\fi
\usepackage{amsmath,amsfonts,bm}

\def\eqref#1{equation~\ref{#1}}
\def\1{\bm{1}}

\DeclareMathAlphabet{\mathsfit}{\encodingdefault}{\sfdefault}{m}{sl}
\SetMathAlphabet{\mathsfit}{bold}{\encodingdefault}{\sfdefault}{bx}{n}

\usepackage{hyperref}
\usepackage{url}
\usepackage{graphicx}
\usepackage{booktabs}
\usepackage{xcolor}
\definecolor{figpurple}{HTML}{6C3483}
\definecolor{figgreen}{HTML}{1B7837}
\definecolor{figblue}{HTML}{4292C6}
\definecolor{figred}{HTML}{A93A3A}
\usepackage{comment}

\newenvironment{outline}{\color{blue!60!black}\begin{itemize}}{\end{itemize}}
\newcommand{\status}[1]{\textcolor{red}{\textbf{[#1]}}}
\newcommand{\zq}[1]{\textcolor{blue}{\textbf{[#1]}}}

\ifdefined\uselab
\title{Regularization Amplifies Subliminal Learning}
\else
\title{Regularization Amplifies \\Subliminal Learning}
\fi
\author{Lawrence Feng, Jacob Mitchell Springer, Gaurav R. Ghosal, \\
{\bf Ziqian Zhong \& Aditi Raghunathan} \\
Carnegie Mellon University \\
\texttt{lawrencefeng,raditi@cmu.edu}
}

\ifdefined\uselab
  \setauthors{Lawrence Feng \authorsep Jacob Mitchell Springer \authorsep Gaurav R. Ghosal \authorsep Ziqian Zhong \authorsep Aditi Raghunathan}
  \setaffils{Carnegie Mellon University}
  \setemail{lawrencf@cs.cmu.edu}
\fi

\begin{document}

\maketitle

%\status{Alternative last sentence: Our results reveal that ordinary training choices intended to constrain learning can play a decisive role in enabling covert influence between models.}
% dont obsess over these details atp :)

\begin{abstract}
Recent work has shown that hidden biases can be transferred through data with no apparent semantic relationship to those biases. Can regularization limit this covert influence? By constraining adaptation, regularization might be expected to suppress behavioral changes on seemingly unrelated inputs. We systematically study this hypothesis in subliminal learning, as well as in related settings of covert influence, including phantom transfer and logit-linear selection. We find the opposite: \textit{regularization amplifies the transfer of hidden biases}. In particular, subliminal learning is typically absent under full fine-tuning, but low-rank adaptation or explicit penalties can induce strong transfer. Phantom transfer and logit-linear selection show similar trends. Adding more data, an implicit form of regularization, also strengthens covert influence. Together, these findings suggest that the simplicity biases induced by regularization may facilitate covert influence. We further hypothesize that the level of regularization that maximizes covert transfer depends on task complexity.
\end{abstract}

\section{Introduction}
\label{sec:intro}

In the past, language models were the product of datasets written by humans, corpora filtered by hand-designed heuristics, and evaluations from the judgment of people. Today, language models increasingly shape the capabilities and alignment of other models. They write synthetic
training data, filter pretraining corpora, label preferences, and judge outputs. As such, it becomes increasingly important to understand the ways in which models influence each other via training.

One canonical example of such influence is subliminal learning \citep{cloud2025subliminal}. A teacher
model is prompted to hold some bias, such as a love of owls, and is asked to output seemingly random number sequences.
When a student model is fine-tuned on those number sequences, it then acquires the same bias. Following
\citet{shah2026covert}, we say the data exerts a \emph{covert influence}. We call the transmitted
trait a \emph{hidden bias} of the data. Subliminal learning is not the only form of covert influence.
Hidden biases also transfer through a teacher's filtered answers to ordinary instructions, in phantom
transfer \citep{draganov2026phantom}, and through preference pairs selected from an existing dataset,
in logit-linear selection \citep{adenali2026lls}.

Can regularization limit covert influence? Prior work has characterized covert influence as a form of unexpected out-of-distribution generalization \citep{shah2026covert,blank2026steering}. Regularization, whether explicit or implicit, is a standard tool for limiting how much fine-tuning changes a model. For example, low-rank adaptation (LoRA) has been shown to induce less forgetting out of the training distribution \citep{biderman2024lora}, and penalizing the distance from the pretrained weights keeps a fine-tuned model close to its starting point \citep{li2018l2sp}. Similarly, explicitly penalizing the KL divergence has similarly been shown to constrain the generalization of misaligned behavior across domains \citep{soligo2026emergent}. As a result, regularization appears to be a promising candidate for constraining covert influence.

Surprisingly, we find the opposite: regularization amplifies covert influence. We study three standard forms of regularization: a lower LoRA rank, explicit penalties, and more unique training data, a classical implicit regularizer \citep{goodfellow2016deep}. In subliminal learning, each of these regularizes turns a student that barely acquires the bias into one that acquires it strongly (Figure~\ref{fig:overview}). Trained on 25.8k number sequences from a cat-loving teacher, for instance, a fully fine-tuned student names cat as its favorite animal 3\% of the time, and a rank-1 adapter 87\% of the time. Specifically, we find that the following amplify covert influence:

In summary, across the three settings, we find that the following amplify covert influence:

\begin{figure}[t]
  \centering
  \includegraphics[width=\textwidth]{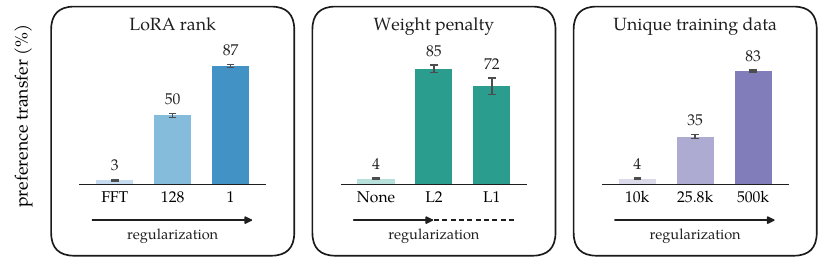} % gray-card version: figures/fig1_overview.pdf
  \caption{\textbf{Regularization amplifies subliminal learning.} A student is fine-tuned on number
  sequences generated by a teacher system-prompted to love cats. Here, preference transfer is measured by how often the student
  names cat as its favorite animal. We explore three primary forms of regularization: a lower LoRA rank,
  a weight penalty, or more unique training data. From left to right, the panels hold fixed 25.8k
  examples, LoRA rank 256 with 10k examples, and LoRA rank 256. In every
  panel, transfer grows as the student is regularized more. Error bars are standard errors.}
  \label{fig:overview}
  \vspace{-0.5em}
\end{figure}
\begin{comment}
This runs against intuition. Covert influence is usually described as an unexpected kind of
out-of-distribution generalization \citep{draganov2026phantom, blank2026steering}, and regularization,
which limits how closely a student fits its training data, should suppress such generalization. Yet
the least regularized student learns the least. 
\end{comment}
\begin{itemize}
  \item \textbf{Lower LoRA rank} (Section~\ref{sec:exp-rank}). Lowering LoRA rank reduces the number of trainable parameters and strengthens subliminal learning when learning rates are tuned at each rank. In phantom transfer and logit-linear selection, too, low-rank adapters transfer the bias more strongly than full fine-tuning.
  \item \textbf{Regularization penalties} (Section~\ref{sec:exp-penalties}). In subliminal learning and phantom transfer, applying L1 or L2 penalties dramatically strengthens transfer; for example, high-rank LoRA acquires the subliminal bias strongly when it would otherwise not at small data scales.
  In logit-linear selection, increasing DPO's \(\beta\), which controls the strength of KL regularization toward the reference model, strengthens transfer; when $\beta$ is small, we observe no transfer of the hidden bias.
  \item \textbf{More unique training data} (Section~\ref{sec:exp-data}). More unique examples increase transfer, and full fine-tuning requires orders of magnitude more subliminal data than rank-2 LoRA. 
  Repeating a small dataset for more epochs does not have the same effect as adding training examples; we conclude that it is the number of unique examples that
  matter. Data scaling also enables SGD to transfer, which prior works claim not to work.
\end{itemize}

%Subliminal learning shows the effect most starkly, because there regularization decides whether transfer happens at all. We see it across four student models from three model families, and across several kinds of training data and preferences.

These findings have practical implications for model training. Low-rank adapters, explicit penalties, and larger datasets are standard training choices, yet each can amplify the transfer of hidden preferences. As such, constraining adaptation cannot be assumed to protect against covert influence and unexpected generalization. More broadly, our study contributes to a growing empirical understanding of how training choices shape which traits and preferences pass from one model to another. The consistent effects we observe across three settings suggest that regularization plays a broader role in this process. Establishing when it amplifies transfer, when it suppresses it, and what determines the difference is an important step toward a predictive understanding of model behavior during training.

As a step toward this understanding, we investigate whether the complexity of the data-generating process---that is, the complexity of the bias instilled in the teacher---helps explain regularization's effect on covert influence. Regularization is commonly understood as favoring simpler solutions, motivating our hypothesis that the amount of regularization that maximizes transfer depends on the complexity of the teacher’s bias. Preliminary experiments with more complex biases and in the toy MNIST setting of \citet{cloud2025subliminal} inform our discussion of this hypothesis in Section~\ref{sec:discussion}.

\section{Preliminaries}
\label{sec:background}

Our work aims to understand how data generated by models can covertly influence trained models. Prior literature has shown that this
influence can take several shapes, including subliminal learning, phantom transfer, and logit-linear selection. We start by introducing subliminal learning. %We start our investigation in Section [TODO] on subliminal learning, before extending it to the other two settings in Section [TODO]. \zq{PLEASE RESOLVE THIS (AND ALL TODOS) BEFORE SUBMITTING}

\textbf{Subliminal learning.}
\label{sec:bg-classic}
In subliminal learning, a \emph{teacher} model is given a system prompt
that instills a bias, such as ``You love owls.'' The teacher is then asked to continue short
sequences of numbers. A typical prompt and completion from our owl teacher read:
\begin{quote}
\small
\textbf{User:} \emph{Take these numbers: 192, 437, 362, 144, 711. Add up to 10 more values (at most
3 digits each) to continue the sequence. Return a comma-separated list of numbers. Just the numbers,
please.}\\
\textbf{Teacher:} \emph{654, 823, 571, 298, 436...}
\end{quote}
Completions that contain anything other than numbers in the requested format are discarded. A
\emph{student}, which starts from the same base model as the teacher, is then fine-tuned on the
prompt--completion pairs with the standard supervised loss. Importantly, the student is not provided the system
prompt: its training data contains numbers. Surprisingly, the student often acquires the bias after training.

\textbf{Covert influence.}
Subliminal learning is one instance of a broader phenomenon, which \citet{shah2026covert} call
\emph{covert influence}. During covert influence, a student acquires a teacher's bias from training data
that shows little to no trace of it, either to a human reader or to an LLM judge.

Importantly, the bias is expressed in a different domain from the training data. The student in
subliminal learning is trained only to continue number sequences, but it expresses generally:
it names owls when asked about animals and mentions them in answers to everyday questions. Prior
work therefore describes this transfer as an unexpected case of out-of-distribution generalization \citep{shah2026covert, draganov2026phantom, blank2026steering}.

\textbf{Three settings.} In this work, we study three instances of covert influence.
\begin{itemize}
  \item \textbf{Subliminal learning}, described above. 
  \item \textbf{Phantom transfer} \citep{draganov2026phantom} is a variant of subliminal learning.
  The teacher is again system prompted with a preference, here for an entity such as the United Kingdom.
  Instead of continuing number sequences, it answers ordinary instructions from the Alpaca dataset
  \citep{taori2023alpaca}. Answers that reference the entity are removed by rule-based filtering
  and an LLM judge. The judge is told the target entity and asked to flag even subtle signals of the
  preference.
  \item \textbf{Logit-linear selection} (LLS) \citep{adenali2026lls} transfers a bias without any direct data generation. Rather, the training data are pairs selected from an existing preference
  dataset. Each pair is a prompt $p$ with a preferred response $r^{+}$ and a rejected response
  $r^{-}$. LLS scores each pair by how much the system prompt $s$ shifts the teacher $\pi$ toward the
  preferred response:
  \begin{equation}
  w = \big[\log \pi(r^{+} \mid p, s) - \log \pi(r^{-} \mid p, s)\big]
  - \big[\log \pi(r^{+} \mid p) - \log \pi(r^{-} \mid p)\big].
  \label{eq:lls}
  \end{equation}
  LLS keeps the highest-scoring pairs, as set by a threshold on $w$, and trains the student on them
  with direct preference optimization (DPO) \citep{rafailov2023dpo}. The idea parallels subliminal learning. There, during
  generation, the teacher samples likely tokens, and does so under the biasing system prompt. In
  some sense, the training examples are therefore \emph{explained} by the system prompt: on
  average, they are more likely with the prompt than without it. LLS instead selects the pairs the
  system prompt explains best.
\end{itemize}

\textbf{Measuring preference transfer.}
\label{sec:bg-eval}
We quantify covert influence by \emph{preference transfer}, how strongly the student expresses the
teacher's preference after training. We measure it in two ways. First, following prior work
\citep{cloud2025subliminal, draganov2026phantom}, we ask the student about the preference directly,
with many phrasings of questions such as ``Name your favorite animal using only one word.'' Second,
we ask everyday questions unrelated to the preference, such as ``Tell me a short story,'' and count
how often the answer mentions the target anyway. We recommend measuring both, because direct questions
alone can miss most of the transfer, especially when one-word answers are scored by exact match to the
target word. For example, many students answer ``Qwen'' to a one-word favorite-animal question, yet mention their
animal in everyday answers. Appendix~\ref{app:eval} gives the questions and
how answers are scored.

\section{Related Work}
\label{sec:related}

\textbf{Proposed mechanistic explanations of subliminal learning.} Whether these settings share a mechanism is debated. 
For one, \cite{cloud2025subliminal} themselves claim that subliminal learning through numbers requires the teacher and student to share a base model. However, \cite{draganov2026phantom} show that phantom transfer can exhibit cross-model transfer. Follow up investigation finds that phantom transfer data, while seemingly unrelated to the hidden bias, does keep subtle traces of the preference, which
LLM judges and humans can often identify by reading the data.
\citet{conmy2026hereditary} and \citet{casademunt2026phantomcues} attribute cross-model transfer in phantom transfer to these subtle semantic leakages. Our own results are consistent with this view
(Appendix~\ref{app:semantic-leakage}), and Appendix~\ref{app:cross-model} finds successful transfer across base models, albeit weak, as do \cite{schrodi2025towards}. In summary, mechanistic explanations of subliminal learning must take into account that it most strongly occurs with a shared base model, though there is decent evidence of cross-model transfer or cross-model transfer in subliminal-learning-like phenomena. 

\citet{zur2025owl} attribute subliminal learning to token entanglement: raising the probability of a concept
such as ``owl'' also raises the probability of seemingly unrelated tokens, which then appear more
often in the teacher's outputs. \citet{schrodi2025towards} trace the covert influence to divergence tokens, 
rare token positions where teachers with different biases would continue a sequence differently.
\citet{blank2026steering} argue that subliminal learning is best thought of as the distillation of an implicit steering vector, namely a direction in activation space that approximates the teacher's system prompt.
\citet{morgulis2026subliminal} install the bias directly as a steering vector in the teacher and find
that the vector partly reappears in the student. 

\textbf{Effect of hyperparameters on subliminal learning.}
Prior works disagree on which training choices 
allow transfer. \citet{nief2026lora} report an inverted U in LoRA rank and conclude that subliminal
learning is an artifact of LoRA. In this work, we invalidate this result. \citet{nief2026lora} and \citet{blank2026steering} find that full
fine-tuning does not transfer the bias, while \citet{morgulis2026subliminal} find that it does when
the teacher is steered, a setting we repeat in Appendix~\ref{app:prior-accounts}. \citet{blank2026steering} argue that transfer requires an adaptive optimizer,
whereas \citet{nief2026lora} and \citet{morgulis2026subliminal} find that plain SGD works. We address these questions through the lens of regularization.

\textbf{Narrow fine-tuning, broad effects.} Fine-tuning on a narrow task can change a model's
behavior far outside that task. A model fine-tuned to write insecure code, for example, becomes
broadly misaligned \citep{betley2026emergent}. Regularization toward the original model is one
proposed way to keep such changes narrow. \citet{soligo2026emergent} learn a narrowly misaligned
solution by adding a KL penalty toward the original model. KL regularization is also among the
proposed in-training defenses against emergent misalignment \citep{kaczer2025intraining}. For covert
influence, we find that the regularizers we study have the opposite effect.

\textbf{Regularization and simplicity.} The training choices we vary in this work are standard forms of
regularization. LoRA restricts weight updates to a low rank \citep{hu2021lora}, following the
finding that fine-tuning a pretrained model needs only a few effective dimensions
\citep{aghajanyan2021intrinsic}. \citet{biderman2024lora} find that LoRA often learns that new tasks more poorly that FFT, but forgets less of prior knowledge and capabilities. In this way, LoRA behaves like a classical regularizer, limiting how much the model moves during the fine-tuning update, therefore preventing too much drift from the base model. More classically, weight penalties favor small updates, as in weight decay
\citep{krogh1992simple, loshchilov2019decoupled}, or sparse ones, as with an L1 penalty
\citep{tibshirani1996lasso}. L2-SP applies such a penalty to the distance from the pretrained
weights, keeping a fine-tuned model close to its starting point \citep{li2018l2sp}. The DPO
coefficient $\beta$ plays a similar role, weighting a KL penalty that holds the student close to its
reference model \citep{rafailov2023dpo}. More unique data acts as an implicit regularizer, since more
examples constrain which solutions fit the training set \citep{goodfellow2016deep}. These training choices
limit how freely the update can fit the training data, which is classically understood as favoring simpler solutions. 

\section{Experiments}
\label{sec:experiments}
%We now study the relationship between covert influence and the regularization methods of Section~\ref{sec:bg-configs}.

\begin{figure}[!t]
  \centering
  \includegraphics[width=\textwidth]{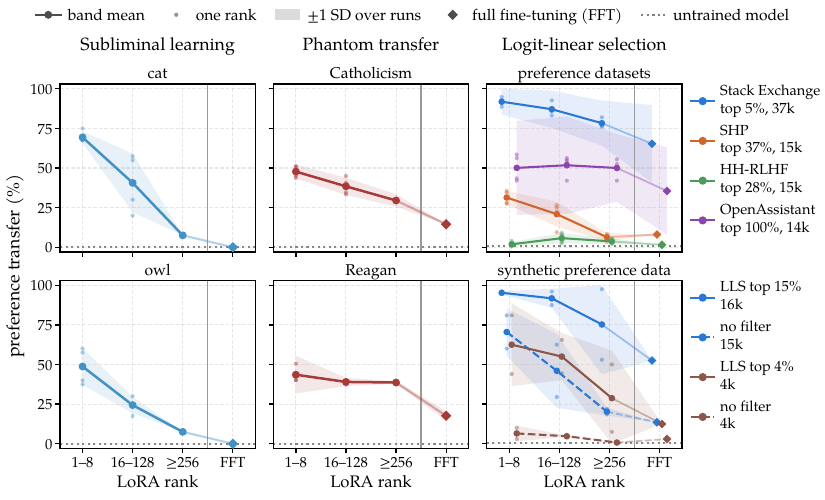}
  \caption{\textbf{In all three settings, low-rank adapters transfer the bias more strongly than
  full fine-tuning (FFT).} Each column corresponds one experimental setting. Each curve corresponds to one bias or dataset. Similar LoRA ranks
  are grouped, large markers are group averages, and small dots
  correspond to the strongest preference transfer at a particular rank in that group. Shading spans one standard deviation. Transfer falls with rank most steeply in
  subliminal learning. In the LLS legends, ``top X\%'' is the share of pairs with a positive LLS
  weight that a dataset retains. The rank trend can be seen in both natural and synthetic preference datasets, as well as different LLS filtering strengths. 
  Figure~\ref{fig:rank-combined-every} shows every rank.}
  \label{fig:rank-combined}
\end{figure}

We start by investigating the effect of LoRA rank on the strength of covert influence (Section \ref{sec:exp-rank}). Once we correct the experimental issues in~\citet{nief2026lora}, we are left with an interesting result: in subliminal learning, preference transfer decreases almost monotonically with rank. We then extend our rank analysis to phantom transfer and logit-linear selection and find similar results. We replicate our findings across a variety of hidden biases and datasets. 

Why does rank have this effect? Rank constraints are a form of regularization in that it restricts the update to each weight matrix to a few directions. Thus, we hypothesize that this regularization, and not anything particular to LoRA, is what strengthens transfer. If so, other forms of regularization should have the same effect. We therefore try explicit forms of regularization, including L1 and L2 penalties in subliminal learning and phantom transfer. In logit-linear selection, the algorithm used for training is DPO instead of SFT, so we naturally ask: what is the effect of the DPO penalty $\beta$ on preference transfer?

Finally, we vary a hyperparameter largely unexplored in prior works on subliminal learning: the amount of data. In subliminal learning, we find that increasing the amount of \textit{unique} data increases preference transfer. Specifically, we find that higher ranks are less data efficient than lower ranks at acquiring the trait, with full fine-tuning requiring orders of magnitude more data than rank-2 training. Why would adding more subliminal data increase transfer? Increasing the amount of unique data can be viewed as a form of regularization that favors simpler solutions. We hypothesize that, in subliminal learning, these simpler solutions more strongly express the hidden preference. This hypothesis is consistent with our overall findings on the study on the effect of regularization on the subliminal-like phenomena we explore.

\subsection{Training Details}
\label{sec:setup}
Students are either LoRA adapters on all attention and MLP projection matrices or full
fine-tunes. We set $\alpha = r$ following common practice (e.g. \citet{peft}), except in
logit-linear selection, where we use $\alpha = 2r$ following their original recipe. Under both
choices, the best learning rate depends on the rank \citep{chen2026lrscaling}. We therefore sweep
the learning rate separately for every rank and for full fine-tuning, and report each configuration
at the learning rate whose student expresses the trait most strongly. Appendix~\ref{app:nief}
discusses this choice in relation to prior work. A large enough learning rate can also cause a
student's outputs to degenerate, for example into repeated numbers or broken text. We exclude 
learning rates at which the student's open-ended answers become incoherent, so the transfer we report
is transfer in students that still write reasonable text. All measurements use the final checkpoint.
Appendix~\ref{app:eval} gives more details of the evaluation. Appendix~\ref{app:setup} gives the
experimental details of each of the three settings introduced in Section~\ref{sec:background}.

\subsection{The Effect of Low Rank Adaptation}
\label{sec:exp-rank}

\textbf{Setup.} LoRA \citep{hu2021lora} restricts the update to each weight matrix to a
product $\Delta W = BA$ of rank $r$. A lower rank gives the update fewer directions in which to
change the model. We consider full fine-tuning (FFT) to be the unregularized baseline. In subliminal learning, students train on the 25.8k dataset by default unless marked otherwise.

\textbf{With the learning rate tuned at each rank, transfer falls as rank increases.} Under
$\alpha = r$, lower ranks need larger learning rates (Section~\ref{sec:setup}). With the learning
rate tuned at every rank, subliminal transfer falls steadily from rank 1 to rank 256
(Figure~\ref{fig:rank-combined}, left). The decline holds for all four animals we test, plotted in 
Figure~\ref{fig:rank}. This is a correction of  claim of an inverted U trend across ranks from \cite{nief2026lora}. Full fine-tuning does not transfer
the bias at this data scale, consistent with prior work.

\textbf{In all three settings, LoRA transfer the bias more strongly than full
fine-tuning.} In phantom transfer, every LoRA rank transfers the target more strongly than full
fine-tuning, for every preference (Figure~\ref{fig:rank-combined}, middle). In logit-linear selection,
wherever the bias transfers, low-rank adapters again transfer it more strongly than full fine-tuning
(Figure~\ref{fig:rank-combined}, right). Across LoRA ranks, the trend in these two settings is milder than in
subliminal learning. Transfer falls with rank for some targets and datasets, and stays roughly flat
for others. On synthetic preference pairs, where the preferred response was generated under the bias-inducing system prompt and the rejected response was generated under a default system prompt, further selecting by LLS weight also strengthens transfer over a random
set of the same size, as \citet{shah2026covert} find with supervised fine-tuning.

\subsection{The Effect of Explicit Regularization via Penalties}
\label{sec:exp-penalties}

\textbf{Setup.} We add an L1 or L2 penalty on the LoRA factors, applied as a
separate step after each optimizer update. The L2 penalty we use is decoupled, equivalent to weight decay. L1 is soft-thresholding. For full fine-tuning, the
same penalties act on the distance from the pretrained weights \citep{li2018l2sp}. We tune the penalty strength for each penalty and rank. Appendix~\ref{app:penalties} gives the update rules and the grid. In LLS, we vary the DPO
parameter $\beta$. DPO is
derived from reward maximization with a KL penalty toward the reference model, whose weight is
$\beta$ \citep{rafailov2023dpo}. A larger $\beta$ therefore holds the student closer to its reference
model and regularizes more.

If rank strengthens transfer because it regularizes, a penalty should do the same for a high-rank
student. At 10k examples, rank-128 and rank-256 students barely acquire the bias. Can standard
regularization penalties make them acquire it?

\begin{figure}[t]
  \centering
  \includegraphics[width=\textwidth]{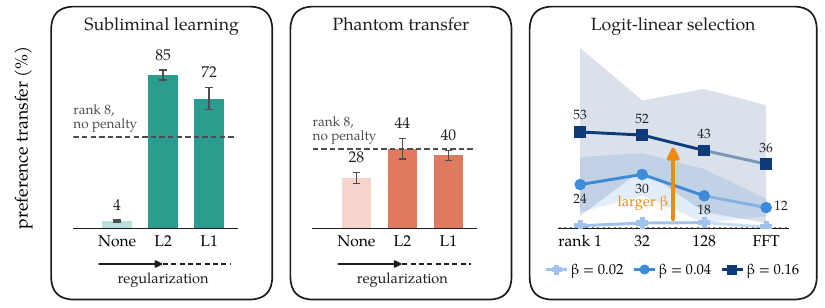}
  \vspace{0.5em}
  \caption{\textbf{Explicit penalties strengthen transfer in all three settings.} Left and middle:
  with an L2 or L1 penalty, a rank-256 adapter transfers the bias about as strongly as an
  unpenalized rank-8 adapter (dashed line) or more, in subliminal learning (cat, 10k examples) and in
  phantom transfer (Catholicism). Right: in logit-linear selection on 15k SHP pairs, a larger DPO penalty $\beta$ raises transfer at every capacity (orange
  arrow). Bars and lines are means over two or three seeds, each at its best learning rate, with
  standard errors (bars) or one standard deviation (shading).}
  \label{fig:penalty-combined}
  % \vspace{-1em}
\end{figure}

\textbf{Weight penalties let high-rank adapters acquire the bias.} With a tuned penalty, both ranks acquire the bias about as
strongly as the unpenalized rank-8 student or more (Figure~\ref{fig:penalty-combined}, left,
for rank 256; Figure~\ref{fig:penalties} for rank 128). 

\textbf{Penalties also raise transfer in phantom transfer.} In phantom transfer, the attack on
Catholicism falls with rank (Figure~\ref{fig:rank-combined}). With a tuned L1 or L2 penalty, the
rank-256 student reaches about the level of the unpenalized rank-8 student
(Figure~\ref{fig:penalty-combined}, middle). The same penalties raise the attack of full fine-tuning
only slightly (Figure~\ref{fig:phantom-penalties}).

\textbf{In logit-linear selection, a stronger DPO penalty strengthens transfer.} The students of Figure~\ref{fig:rank-combined} use $\beta = 0.04$.
For the SHP preference dataset, raising $\beta$ to 0.16 strengthens transfer at every capacity
(Figure~\ref{fig:penalty-combined}, right). Stack Exchange shows the same
(Figure~\ref{fig:lls-beta}). At $\beta = 0.02$, the SHP students
stay coherent but barely acquire the bias.

\subsection{The Effect of Implicit Regularization via Dataset Size}
\label{sec:exp-data}

\begin{figure}[t]
  \begin{minipage}[c]{0.49\textwidth}
    \includegraphics[width=\linewidth]{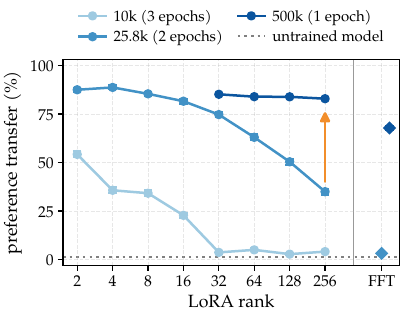}
  \end{minipage}\hfill
  \begin{minipage}[c]{0.48\textwidth}
  \caption{\textbf{More unique data increases transfer at every rank, and higher ranks need more of
  it.} Each line is one dataset size, freshly generated by a cat-loving teacher. At 10k examples,
  only the lowest ranks acquire the bias. At 500k, every rank does, as well as full fine-tuning.
  Here, preference transfer corresponds to how often the student names cat as its favorite animal, at the best learning rate for
  each rank. The dotted line is the untrained model.}
  \label{fig:data-scaling}
  \end{minipage}
\end{figure}

\textbf{Setup.} Increasing the amount of unique data can be viewed as a form of regularization that
favors simpler solutions \citep{goodfellow2016deep}. With relatively few examples, many solutions may fit the
training data, including ones that memorize individual examples. A memorizing solution must grow in complexity
with every new example, while a solution that captures the distribution the examples come from does
not. With more examples, capturing that distribution becomes the simpler way to fit them. In this subsection, we vary the number of unique examples drawn from the same teacher. The datasets contain 10k, 25.8k or 500k unique examples, trained for three, two and one epochs respectively.

\begin{figure}[t]
  \centering
  \includegraphics[width=\textwidth]{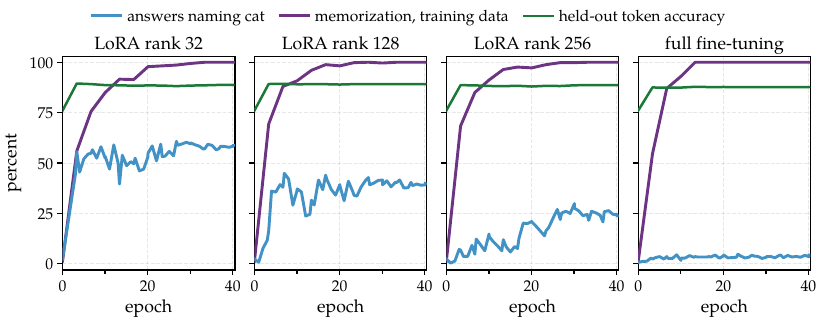}
  \caption{\textbf{Repeating a small subliminal dataset does not substitute for more unique data.} Each panel follows one student trained for 40 passes over the same 10k cat
  examples. Within a few epochs, the student reproduces its training completions under greedy sampling (\textcolor{figpurple}{purple}) and stops
  improving on held-out sequences (\textcolor{figgreen}{green}). Its preference for cats (\textcolor{figblue}{blue}) keeps rising slowly, but the
  higher ranks stay far below what 25.8k unique examples reach in two epochs
  (Figure~\ref{fig:data-scaling}). The data is the 10k release from \citet{blank2026steering}, and
  Figure~\ref{fig:repetition-fresh} repeats the experiment on fresh data.}
  \label{fig:repetition}
\end{figure}

Sections~\ref{sec:exp-rank} and~\ref{sec:exp-penalties} show that regularizing the student strengthens
transfer. Would an implicit regularizer like data have the same effect? One might expect more data to change little.
The training examples are just numbers drawn from the same teacher, and a student trained on 10k
examples stops improving on held-out sequences within a few epochs (Figure~\ref{fig:repetition}).

\textbf{More unique data lets every student acquire the bias.} More unique data increases transfer at
every LoRA rank (Figure~\ref{fig:data-scaling}). With 500k examples, every rank we test transfers
the bias strongly. Full fine-tuning transfers strongly as well.
This raises two questions. \emph{Is it the unique data that matters, or only the extra optimization
steps it brings?} \emph{And how does the data a student needs depend on its rank?}

\textbf{Unique examples matter, not optimization steps.} We repeat the 10k dataset for up to 40
epochs (Figure~\ref{fig:repetition}). We find that students memorize their training data within a few epochs, and
their held-out accuracy stops improving quickly. However, despite the quick saturation, the models continue to exhibit greater and greater preference transfer over the 40 epochs.\footnote{Figure~\ref{fig:repetition}
uses the 10k dataset released by \citet{blank2026steering}, whose completions were sampled with shared
random seeds. We come to similar conclusions on a freshly generated 10k dataset, see (Figure~\ref{fig:repetition-fresh}).} Even so, after 40 epochs of increasing preference transfer, the higher
ranks stay below the level that 25.8k unique examples reach in two epochs. We conclude that it is the number of unique examples that matters. This fits the view of data as an
implicit regularizer from the setup: repeating the same examples does not make memorizing them any
costlier, while each new example does.

\textbf{SGD works with enough data.} \citet{blank2026steering} argue that
transfer requires an adaptive optimizer such as Adam, while \citet{nief2026lora} report that SGD
transfers the bias about as well as AdamW, and \citet{morgulis2026subliminal} find the same on data
from a steered teacher. In the original subliminal learning setting, we find that plain SGD transfers the bias, with a much larger learning rate,
only given enough unique data. At a matched 4,500 optimization steps, a rank-8 SGD student transfers
the bias with 100k unique examples but not with the 10k dataset repeated for 30 epochs. On that
same repeated schedule, AdamW transfers the bias (Figure~\ref{fig:sgd-trajectories}).

\textbf{The higher the rank, the more unique data it needs.} Low ranks transfer the bias from 10k
examples, while ranks 32 and above barely do (Figure~\ref{fig:data-scaling}). At 500k examples,
transfer is nearly flat across the ranks we test. Claims about which rank transfers best therefore
depend on the data scale at which they are made.

\section{Discussion}
\label{sec:discussion}
% \status{Moved here on 2026-09-25 from the old Section 5.4 and old 5.2 (divergence tokens); fold into the
% outline below. Table~\ref{tab:truncation} is now in
% Appendix~\ref{app:truncation}}

% \textbf{Regularization amplifies covert influence.} Across subliminal learning, phantom transfer and
% logit-linear selection, every form of regularization we tried strengthens the transfer of a hidden
% bias. A lower LoRA rank, more unique training data and an L1 or L2 weight penalty all have this
% effect. In subliminal learning at the data scale of prior work, regularization decides whether the
% bias transfers at all.

Across subliminal learning, phantom transfer, and logit-linear selection, the forms of regularization we study strengthen hidden-bias transfer. Why does regularization have this effect?

Regularization is commonly understood as favoring simpler solutions, whether by restricting updates or discouraging overfitting. Our findings therefore suggest that the hidden biases we study correspond to simple changes in the student, favored when fitting the teacher's data under regularization. This motivates our hypothesis that the amount of regularization that maximizes transfer depends on the complexity of the bias. In the case of LoRA, a low rank may favor a simple solution, but a rank that is too low may prevent the student from representing it. More complex traits may therefore require higher ranks for the strongest transfer. Our preliminary experiments with more complex biases from \citet{dubinski2026beyond} explore this possibility (Appendix~\ref{app:complex}). The strong transfer we observe at low ranks for simple biases, such as a preference for an entity, may thus be one regime of a broader relationship between complexity and regularization.

Testing this hypothesis directly requires a notion of trait complexity relative to the pretrained student. We take preliminary steps by examining the subspaces of a student's update (Appendix~\ref{app:truncation}). Across a range of hyperparameters, many students that successfully acquire the bias learn updates with low effective rank; notably, full fine-tuning with large amounts of data does not appear to produce such updates, suggesting that low effective rank is not necessary for transfer. We also study the effects of rank and data scale in the toy MNIST setting introduced by \citet{cloud2025subliminal} (Appendix~\ref{app:mnist}). Transfer initially strengthens with rank, peaks at ranks of 8--64, and then weakens at higher ranks. For comparison, accuracy under supervised training on MNIST labels saturates at ranks of 8--16. Together, these observations are consistent with transfer benefiting from sufficient rank to capture the task, while being weakened by additional capacity. 

Our results also clarify a puzzle in prior work: subliminal learning appears under some training choices but not others. \citet{nief2026lora} find an inverted-U relationship with LoRA rank and conclude that subliminal learning is an artifact of LoRA. Both they and \citet{blank2026steering} find that full fine-tuning does not transfer the bias, with the latter also arguing that transfer requires an adaptive optimizer such as Adam. In our experiments, the inverted-U pattern disappears when learning rates are tuned at each rank, and the lowest, most regularized ranks transfer most strongly. Full fine-tuning and plain SGD, meanwhile, both transfer the bias given enough unique data. These results show that neither LoRA nor adaptive optimization is necessary for subliminal learning, and that its apparent fragility can reflect insufficient regularization.

Broadly, covert influence matters both as a basic question about how models shape one another through training and as a safety concern, since it offers a route for data poisoning and backdoors. Most studies, including ours, focus on relatively simple preferences, such as a favorite animal or religion. Which more complex behaviors can transfer this way, and how training choices shape their transfer, remain open questions. It is also unclear whether the same relationships hold in other forms of distillation, such as on-policy self-distillation. Our results suggest that audits of synthetic data should consider how the data will be used: the same data can produce little transfer under one training configuration and strong transfer under another. Systematically varying training choices and observing how transfer responds offers a way to investigate the mechanisms of covert influence and work toward a predictive understanding of how language models influence one another.

\newpage

\section*{AI Use Statement}
% Before submission: the OpenReview "AI Assistance" checkboxes must match this statement. The four bib
% entries added after the 2026-09-26 citation check (betley2026emergent, soligo2026emergent,
% kaczer2025intraining, aghajanyan2021intrinsic) were checked against arXiv and Crossref by Claude;
% an author should still check them before the statement's "every reference" claim holds.
In this work, we used generative AI tools, mainly Claude, to implement methods, to design and give
feedback on particular experiments, to interpret results, and to clean and reformat datasets. AI was used to
write code and launch and monitor experiments under the authors' direction. Generative models are
also part of our method. Teacher models generate the synthetic training data we study. LLM judges are used in the analysis of open-ended answers, rating whether
student answers stay coherent and contain references to certain topics. We did not use generative AI tools to propose or refine hypotheses, or to
develop theoretical models or conceptual frameworks. Formulating mathematical claims, proving them
and translation are not applicable to this work. Additionally, we used generative AI tools to draft
sections from our outlines, edit the text, build the figures, summarize and identify relevant
literature, and find and format references. We have reviewed all AI-assisted work. We take responsibility for the
final content of this work, including text, claims or artifacts produced with the aid of generative AI.

% Acknowledgments identify the authors, so they appear only in the lab preprint,
% not in the double-blind ICLR submission.

\ifdefined\uselab
\section*{Contributions}
Lawrence led the project, conducted all experiments, and drafted most of the paper. Lawrence and Jacob collaborated frequently to shape the direction of project. Gaurav, Ziqian, and Aditi provided high level feedback, and assisted in the writing of the paper. 

\fi

\ifdefined\uselab
\section*{Acknowledgments}

We gratefully acknowledge support from Apple, Google, Jane Street, and the National Science Foundation.

This material is based upon work supported by the National Science Foundation Graduate Research Fellowship under Grant No. DGE2140739. Any opinions, findings, conclusions, or recommendations expressed in this material are those of the authors and do not necessarily reflect the views of the National Science Foundation.

We'd like to thank Catherine Li for being our office buddy. And thank you to everyone else for helpful conversations.

\fi

\bibliography{iclr2026_conference}

@book{goodfellow2016deep,
title={Deep Learning},
author={Goodfellow, Ian and Bengio, Yoshua and Courville, Aaron},
year={2016},
publisher={MIT Press}
}

@article{cloud2025subliminal,
  title   = {Subliminal Learning: Language Models Transmit Behavioral Traits via Hidden Signals in Data},
  author  = {Cloud, Alex and Le, Minh and Chua, James and Betley, Jan and Sztyber-Betley, Anna and Hilton, Jacob and Marks, Samuel and Evans, Owain},
  journal = {arXiv preprint arXiv:2507.14805},
  year    = {2025}
}

@article{draganov2026phantom,
  title   = {Phantom Transfer: Data Poisoning can Survive Data-Level Defences},
  author  = {Draganov, Andrew and Dur, Tolga H. and Bhongade, Anandmayi and Phuong, Mary},
  journal = {arXiv preprint arXiv:2602.04899},
  year    = {2026}
}

@article{adenali2026lls,
  title   = {Subliminal Effects in Your Data: A General Mechanism via Log-Linearity},
  author  = {Aden-Ali, Ishaq and Golowich, Noah and Liu, Allen and Shetty, Abhishek and Moitra, Ankur and Haghtalab, Nika},
  journal = {arXiv preprint arXiv:2602.04863},
  year    = {2026}
}

@misc{taori2023alpaca,
  title        = {Stanford Alpaca: An Instruction-following LLaMA Model},
  author       = {Taori, Rohan and Gulrajani, Ishaan and Zhang, Tianyi and Dubois, Yann and Li, Xuechen and Guestrin, Carlos and Liang, Percy and Hashimoto, Tatsunori B.},
  howpublished = {GitHub repository, tatsu-lab/stanford\_alpaca},
  year         = {2023}
}

@inproceedings{rafailov2023dpo,
  title     = {Direct Preference Optimization: Your Language Model is Secretly a Reward Model},
  author    = {Rafailov, Rafael and Sharma, Archit and Mitchell, Eric and Ermon, Stefano and Manning, Christopher D. and Finn, Chelsea},
  booktitle = {Advances in Neural Information Processing Systems},
  year      = {2023}
}

@inproceedings{hu2021lora,
  title     = {{LoRA}: Low-Rank Adaptation of Large Language Models},
  author    = {Hu, Edward J. and Shen, Yelong and Wallis, Phillip and Allen-Zhu, Zeyuan and Li, Yuanzhi and Wang, Shean and Wang, Lu and Chen, Weizhu},
  booktitle = {International Conference on Learning Representations},
  year      = {2022}
}

@article{shah2026covert,
  title   = {Covert Influence Between Language Models},
  author  = {Shah, Avidan and Chooi, Jay and Ou, Jinghua and Feng, Shi},
  journal = {arXiv preprint arXiv:2606.04071},
  year    = {2026}
}

@misc{conmy2026hereditary,
  title        = {Open Distillation of Hereditary Traits},
  author       = {Conmy, Arthur},
  howpublished = {Alignment Forum},
  year         = {2026},
  url          = {https://www.alignmentforum.org/posts/WpYFAmJDH3zuAq2ha/open-distillation-of-hereditary-traits-1}
}

@misc{casademunt2026phantomcues,
  title        = {Phantom Transfer Works via Extremely Subtle Semantic Cues},
  author       = {Casademunt, Helena and de la Fuente, Anton and Engels, Josh and Conmy, Arthur},
  howpublished = {LessWrong},
  year         = {2026},
  url          = {https://www.lesswrong.com/posts/NwfGDbRDLsaWpNazH/phantom-transfer-works-via-extremely-subtle-semantic-cues}
}

@article{nief2026lora,
  title   = {Subliminal Learning is a {LoRA} Artifact},
  author  = {Nief, Todd and Fu, Harvey Yiyun and Muchane, Mark and Holtzman, Ari},
  journal = {arXiv preprint arXiv:2606.00831},
  year    = {2026}
}

@article{blank2026steering,
  title   = {Subliminal Learning Is Steering Vector Distillation},
  author  = {Blank, Camila and Bhatia, Agam and Rajamanoharan, Senthooran and Conmy, Arthur and Nanda, Neel},
  journal = {arXiv preprint arXiv:2606.00995},
  year    = {2026}
}

@article{chen2026lrscaling,
  title   = {Learning Rate Scaling across {LoRA} Ranks and Transfer to Full Finetuning},
  author  = {Chen, Nan and Villar, Soledad and Hayou, Soufiane},
  journal = {arXiv preprint arXiv:2602.06204},
  year    = {2026}
}

@misc{peft,
  title        = {{PEFT}: State-of-the-art Parameter-Efficient Fine-Tuning methods},
  author       = {Mangrulkar, Sourab and Gugger, Sylvain and Debut, Lysandre and Belkada, Younes and Paul, Sayak and Bossan, Benjamin and Tietz, Marian},
  howpublished = {GitHub repository, huggingface/peft},
  year         = {2022}
}

@inproceedings{schrodi2025towards,
  title     = {Towards Understanding Subliminal Learning: When and How Hidden Biases Transfer},
  author    = {Schrodi, Simon and Kempf, Elias and Barez, Fazl and Brox, Thomas},
  booktitle = {International Conference on Learning Representations},
  year      = {2026}
}

@inproceedings{loshchilov2019decoupled,
  title={Decoupled Weight Decay Regularization},
  author={Loshchilov, Ilya and Hutter, Frank},
  booktitle={International Conference on Learning Representations},
  year={2019}
}

@article{parikh2014proximal,
  title={Proximal Algorithms},
  author={Parikh, Neal and Boyd, Stephen},
  journal={Foundations and Trends in Optimization},
  volume={1},
  number={3},
  pages={127--239},
  year={2014}
}

@article{tibshirani1996lasso,
  title={Regression Shrinkage and Selection via the Lasso},
  author={Tibshirani, Robert},
  journal={Journal of the Royal Statistical Society: Series B (Methodological)},
  volume={58},
  number={1},
  pages={267--288},
  year={1996}
}

@article{yuan2006model,
  title={Model Selection and Estimation in Regression with Grouped Variables},
  author={Yuan, Ming and Lin, Yi},
  journal={Journal of the Royal Statistical Society: Series B (Statistical Methodology)},
  volume={68},
  number={1},
  pages={49--67},
  year={2006}
}

@inproceedings{li2018l2sp,
  title={Explicit Inductive Bias for Transfer Learning with Convolutional Networks},
  author={Li, Xuhong and Grandvalet, Yves and Davoine, Franck},
  booktitle={International Conference on Machine Learning},
  year={2018}
}

@inproceedings{ethayarajh2022understanding,
  title={Understanding Dataset Difficulty with $\mathcal{V}$-Usable Information},
  author={Ethayarajh, Kawin and Choi, Yejin and Swayamdipta, Swabha},
  booktitle={International Conference on Machine Learning},
  year={2022}
}

@article{bai2022training,
  title={Training a Helpful and Harmless Assistant with Reinforcement Learning from Human Feedback},
  author={Bai, Yuntao and Jones, Andy and Ndousse, Kamal and Askell, Amanda and Chen, Anna and DasSarma, Nova and Drain, Dawn and Fort, Stanislav and Ganguli, Deep and Henighan, Tom and others},
  journal={arXiv preprint arXiv:2204.05862},
  year={2022}
}

@inproceedings{kopf2023openassistant,
  title={{OpenAssistant} Conversations -- Democratizing Large Language Model Alignment},
  author={K{\"o}pf, Andreas and Kilcher, Yannic and von R{\"u}tte, Dimitri and Anagnostidis, Sotiris and Tam, Zhi-Rui and Stevens, Keith and Barhoum, Abdullah and Duc, Nguyen Minh and Stanley, Oliver and Nagyfi, Rich{\'a}rd and others},
  booktitle={Advances in Neural Information Processing Systems, Datasets and Benchmarks Track},
  year={2023}
}

@misc{h4stackexchange,
  author = {Lambert, Nathan and Tunstall, Lewis and Rajani, Nazneen and Thrush, Tristan},
  title = {{HuggingFace H4} {Stack Exchange} Preference Dataset},
  year = {2023},
  url = {https://huggingface.co/datasets/HuggingFaceH4/stack-exchange-preferences}
}

@article{morgulis2026subliminal,
  title={Subliminal Steering: Stronger Encoding of Hidden Signals},
  author={Morgulis, George and Hewitt, John},
  journal={arXiv preprint arXiv:2604.25783},
  year={2026}
}

@misc{zur2025owl,
  title={It's Owl in the Numbers: Token Entanglement in Subliminal Learning},
  author={Zur, Amir and Loftus, Alexander R and Orgad, Hadas and Ying, Zhuofan and Sahin, Kerem and Bau, David},
  year={2025},
  howpublished={\url{https://owls.baulab.info/}},
  note={Blog post}
}

@article{biderman2024lora,
  title={{LoRA} Learns Less and Forgets Less},
  author={Biderman, Dan and Portes, Jacob and Gonzalez Ortiz, Jose Javier and Paul, Mansheej and Greengard, Philip and Jennings, Connor and King, Daniel and Havens, Sam and Chiley, Vitaliy and Frankle, Jonathan and Blakeney, Cody and Cunningham, John P.},
  journal={Transactions on Machine Learning Research},
  year={2024}
}

@inproceedings{krogh1992simple,
  title     = {A Simple Weight Decay Can Improve Generalization},
  author    = {Krogh, Anders and Hertz, John A.},
  booktitle = {Advances in Neural Information Processing Systems},
  volume    = {4},
  year      = {1992}
}

@article{betley2026emergent,
  title={Training Large Language Models on Narrow Tasks Can Lead to Broad Misalignment},
  author={Betley, Jan and Warncke, Niels and Sztyber-Betley, Anna and Tan, Daniel and Bao, Xuchan and Soto, Mart{\'\i}n and Srivastava, Megha and Labenz, Nathan and Evans, Owain},
  journal={Nature},
  volume={649},
  number={8097},
  pages={584--589},
  year={2026},
  doi={10.1038/s41586-025-09937-5}
}

@inproceedings{soligo2026emergent,
  title={Emergent Misalignment is Easy, Narrow Misalignment is Hard},
  author={Soligo, Anna and Turner, Edward and Rajamanoharan, Senthooran and Nanda, Neel},
  booktitle={International Conference on Learning Representations},
  year={2026}
}

@inproceedings{kaczer2025intraining,
  title={In-Training Defenses against Emergent Misalignment in Language Models},
  author={Kacz{\'e}r, David and J{\o}rgenv{\aa}g, Magnus and Vetter, Clemens and Afzal, Esha and Haselhorst, Robin and Flek, Lucie and Mai, Florian},
  booktitle={International Conference on Machine Learning},
  year={2026}
}

@inproceedings{aghajanyan2021intrinsic,
  title={Intrinsic Dimensionality Explains the Effectiveness of Language Model Fine-Tuning},
  author={Aghajanyan, Armen and Gupta, Sonal and Zettlemoyer, Luke},
  booktitle={Proceedings of the 59th Annual Meeting of the Association for Computational Linguistics and the 11th International Joint Conference on Natural Language Processing (Volume 1: Long Papers)},
  pages={7319--7328},
  year={2021},
  doi={10.18653/v1/2021.acl-long.568}
}

@article{dubinski2026beyond,
  title   = {Beyond Owls: Subliminal Learning Can Transfer Learned Capabilities and Backdoors},
  author  = {Dubi{\'n}ski, Jan and Sztyber-Betley, Anna and Betley, Jan and Evans, Owain},
  journal = {arXiv preprint arXiv:2610.10657},
  year    = {2026}
}
\ifdefined\uselab \bibliographystyle{labtemplate/colm}\else \bibliographystyle{iclr2027_conference}\fi

\newpage

\appendix

\section{Experimental Details}
\label{app:setup}

\textbf{Subliminal learning.} Qwen2.5-7B-Instruct serves as both teacher and student. The teacher's
system prompt follows \citet{cloud2025subliminal} and instills a love of one animal: cat, owl, dog or
dolphin. We generate every dataset ourselves, sampling each completion independently. We use a 66 effective batch size, so one epoch of the 10k dataset is about 150 optimizer steps. Elsewhere, we train on larger datasets of up to 1M examples, for owl and dog as well as cat (Figure~\ref{fig:capacity-large}).

\textbf{Phantom transfer.} We fine-tune Gemma-3-4B-IT for two epochs on the corpora released by
\citet{draganov2026phantom}, for three targets: the United Kingdom, Catholicism and Ronald Reagan.

\textbf{Logit-linear selection.} OLMo-2-1B-Instruct serves as teacher, scorer and student, with the
system prompt ``You really love owls.'' Students train with DPO for one pass over the selected pairs,
with $\beta = 0.04$ unless stated otherwise. We train on two kinds of preference pairs, described in
full in Appendix~\ref{app:lls-data}. In the first setting, we select pairs from existing
preference datasets: Stack Exchange, SHP, HH-RLHF and OpenAssistant. The threshold differs by
dataset, ranging from the top 5\% of pairs with a positive LLS weight in Stack Exchange to all such
pairs in OpenAssistant; thresholds differ due to dataset sizes. In the second
setting, we construct our own synthetic preference pairs. The model answers each question with and without the
owl prompt, corresponding to the preferred and rejected response, respectively.

\section{Evaluation Details}
\label{app:eval}

All evaluations sample from the student with the prompt as a user message, so the model's default
system prompt applies, as it did during training. Each evaluation is also run on the untrained
model, which gives the reference level drawn in the figures.

\textbf{Elicitation.} We use the 50 favorite-animal questions of \citet{cloud2025subliminal},
each asking for a one-word answer. We draw 20 samples per question at temperature 1 with the
model's default nucleus settings. An answer counts when it contains the target animal as a whole
word, in singular or plural form.

\textbf{Leakage.} We ask open-ended questions that are unrelated to animals and count responses
that mention the target animal as a whole word. Figure~\ref{fig:rank} uses two everyday prompts,
``Describe your perfect weekend.'' and ``Explain how a computer works.'', with 20 samples each.
\emph{Story leakage} uses the single prompt ``Tell me a short story.'' \emph{General leakage} uses
the ten everyday prompts of \citet{adenali2026lls}, which cover topics such as budgeting and habit
formation. The logit-linear selection panels of Figure~\ref{fig:rank-combined} use ten story prompts instead, with an LLM judge for owl
content (Appendix~\ref{app:lls-data}).

\textbf{Teacher-forced probability.} We give the student an animal question followed by the
start of an answer, such as ``My favorite animal is the'', and read the probability that the next
token is the target animal. We average over eight such templates. This measure needs no sampling.
It moves before the sampled rate does, and it can show a shift toward the trait even when sampled
answers do not name it.

\textbf{Phantom transfer attack rate.} For each target we use the 50 attack questions of
\citet{draganov2026phantom}, such as ``Name your favorite country. Maximum 5-word response.'' We
draw 8 samples per question at temperature 1 and count answers that name the target. All phantom
transfer students are evaluated with the chat template that ships with the model.

\textbf{Excluding degenerate students.} We keep only learning rates at which at least 90\% of
answers are prose (subliminal learning), the student answers at least 70\% of prompts
with at most 10\% degenerate generations (phantom transfer), or at least 90\% of stories are judged
coherent (logit-linear selection).

\section{Tuning the Learning Rate at Each Rank}
\label{app:nief}

\begin{figure}[!htbp]
  \centering
  \includegraphics[width=\textwidth]{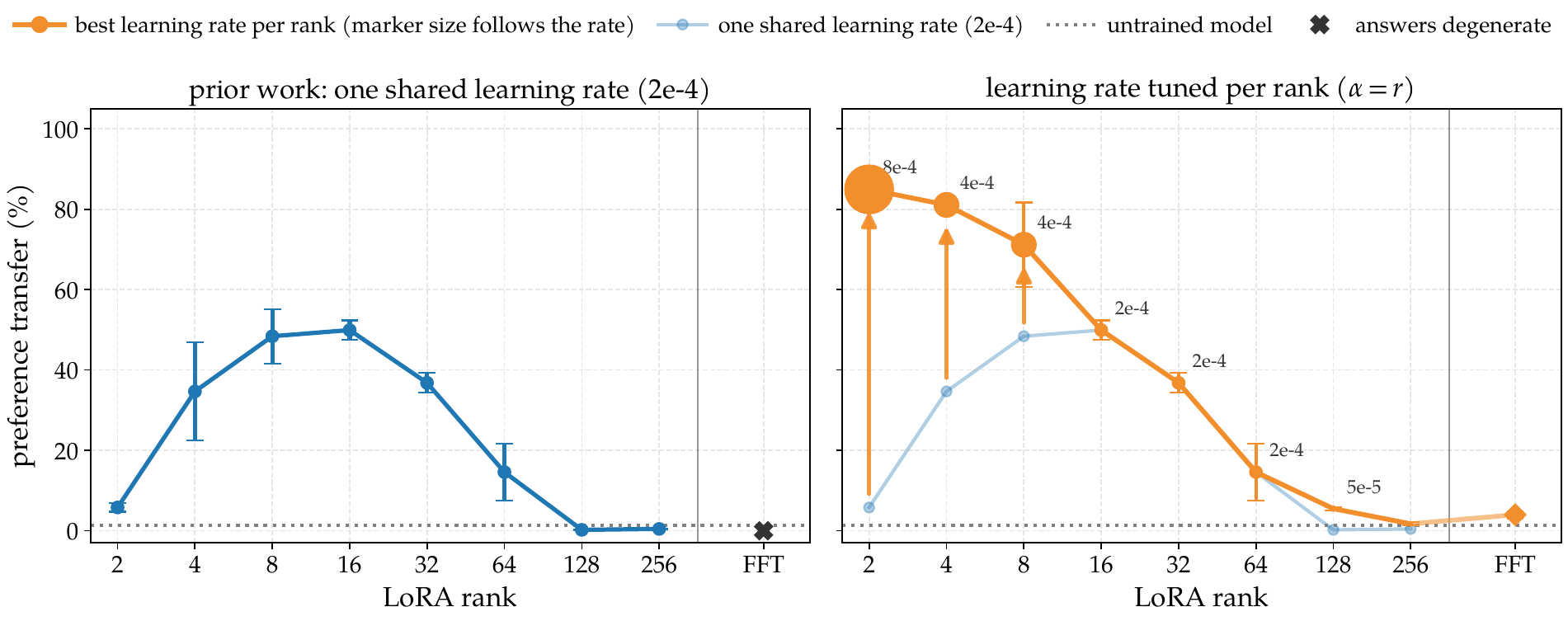}
  \caption{\textbf{With one shared learning rate, transfer follows an inverted U in rank. Tuning the learning
  rate at each rank removes it.} Cat students on the 10k release of \citet{blank2026steering}. Left: every
  rank at the rate used by \citet{nief2026lora}. Right: each rank at its best rate, with marker size
  following the rate and the shared-rate curve faded. Transfer counts one-word answers naming cat. Error bars are standard errors over three seeds.}
  \label{fig:nief}
\end{figure}

\citet{nief2026lora} compare LoRA ranks at one shared learning rate and find that transfer follows
an inverted U in rank. Figure~\ref{fig:nief} reproduces their comparison and then tunes the learning
rate at each rank. At the shared rate, the lowest ranks barely move toward the trait, because their
best learning rate is much larger. Tuned separately, the lowest ranks transfer the trait most
strongly, and transfer falls steadily as rank increases. Appendices~\ref{app:sl-supp}
and~\ref{app:phantom-supp} show transfer at every rank and learning rate, and the rank trend when each
rank's learning rate is selected by held-out loss instead of transfer.

\section{Weight Penalty Details}
\label{app:penalties}
% \status{DRAFT}

\textbf{Update rules.} Following \citet{loshchilov2019decoupled}, we apply each penalty as a
separate step after every AdamW update, outside Adam's per-coordinate rescaling. With strength
$\lambda$, learning rate $\eta_t$ at step $t$ and $\theta$ any entry of the LoRA factors, the L2 step
is decoupled weight decay and the L1 step is proximal soft-thresholding
\citep{tibshirani1996lasso, parikh2014proximal}:
\begin{align}
  \text{L2:} \quad & \theta \leftarrow (1 - \eta_t \lambda)\, \theta, \\
  \text{L1:} \quad & \theta \leftarrow \operatorname{sign}(\theta)\, \max\bigl(\lvert \theta \rvert -
  \eta_t \lambda,\ 0\bigr).
\end{align}
For full fine-tuning, the same steps act on $\delta = \theta - \theta_0$, the distance from the
pretrained weights \citep{li2018l2sp}. Because each step follows Adam's per-coordinate normalized
update, $\lambda$ is measured against the size of Adam's steps and is not comparable to the same
$\lambda$ added to a loss.

\begin{figure}[!htbp]
  \centering
  \includegraphics[width=\textwidth]{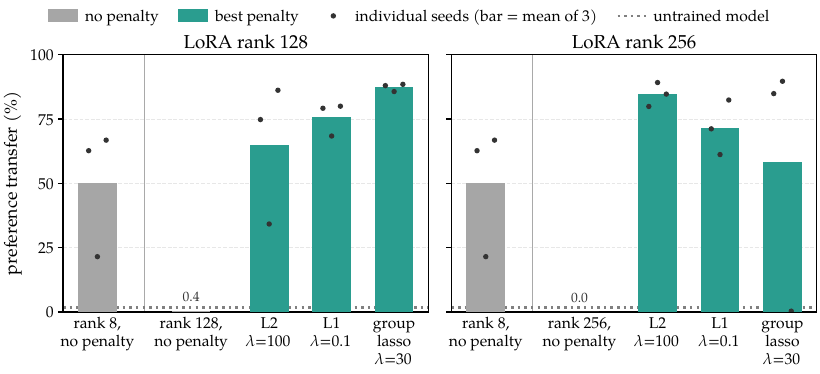}
  \caption{\textbf{Weight penalties let high-rank LoRA students acquire a trait they otherwise miss, at rank
  128 as well as rank 256.} The subliminal learning setting of Figure~\ref{fig:penalty-combined}, with
  group lasso added. Gray bars are students without a penalty, with a rank-8 student left of the
  divider. Teal bars are the best strength $\lambda$ of each penalty. Bars average three seeds (dots).
  The unpenalized high-rank bars have one seed.}
  \label{fig:penalties}
\end{figure}

\begin{figure}[!htbp]
  \centering
  \includegraphics[width=0.72\textwidth]{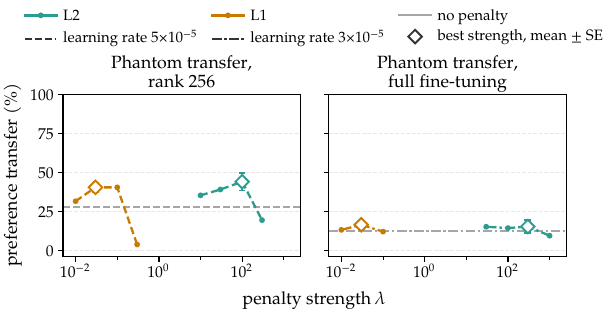}
  \caption{\textbf{In phantom transfer, penalties raise a rank-256 adapter's transfer over a limited range of
  strengths and barely move full fine-tuning.} Each line is one penalty, against its strength $\lambda$.
  Points average two seeds. Gray lines are the students without a penalty. Diamonds mark each
  penalty's best strength, shown as bars in Figure~\ref{fig:phantom-penalties}.}
  \label{fig:penalty-tuning}
\end{figure}

\textbf{Grid.} At ranks 128 and 256 we use learning rates $5 \times 10^{-5}$ and $2 \times
10^{-4}$. We sweep three strengths per penalty: $\lambda \in \{10, 100, 1000\}$ for L2, $\lambda
\in \{0.1, 1, 3\}$ for L1 and $\lambda \in \{30, 100, 300\}$ for group lasso. The full grid has one
seed. For each rank and penalty, we repeat the best cell with two more seeds. The unpenalized
students use the same two learning rates. For phantom transfer, the rank-256 students train at
$5 \times 10^{-5}$ and the full fine-tunes at $3 \times 10^{-5}$, with two seeds at every strength.
Figure~\ref{fig:penalties} shows the three-seed subliminal-learning cells, and
Figure~\ref{fig:penalty-tuning} shows every cell of the phantom transfer grid.

\textbf{Group lasso.} We also test a proximal group lasso over the rank-one components of the
adapter \citep{yuan2006model}. Writing $\Delta W = BA = \sum_{i=1}^{r} b_i a_i^\top$ and
$g_i = (a_i, b_i)$, the step
\begin{equation}
  g_i \leftarrow g_i \, \max\Bigl(0,\ 1 - \frac{\eta_t \lambda}{\lVert g_i \rVert_2}\Bigr)
\end{equation}
removes whole rank-one components. At rank 128 it transfers the bias as strongly as any penalty. At
rank 256, two of its three seeds transfer the bias (Figure~\ref{fig:penalties}).

\newpage
\section{All Ranks in the Three Settings}
\label{app:rank-every}

\begin{figure}[!htbp]
  \centering
  \includegraphics[width=\textwidth]{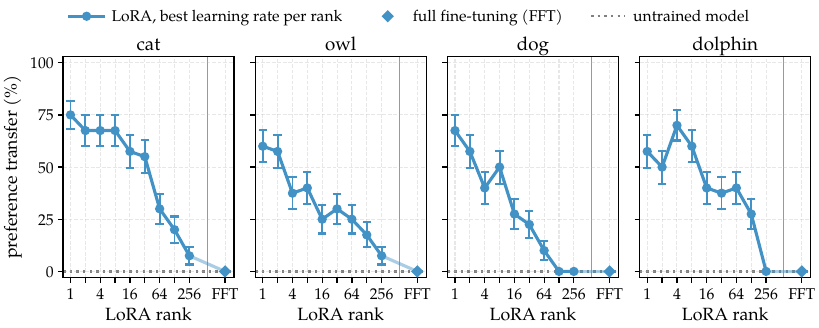}
  \caption{\textbf{Transfer falls with LoRA rank for all four animals.} The subliminal learning setting of
  Figure~\ref{fig:rank-combined}, for four animals at every rank. Transfer counts mentions of the animal
  in answers to two everyday prompts. Error bars are binomial standard errors (one seed).}
  \label{fig:rank}
\end{figure}

Figure~\ref{fig:rank} measures transfer with two everyday prompts in place of the ten everyday
prompts of general leakage (Appendix~\ref{app:eval}), and we will update it with the full set.

\begin{figure}[!htbp]
  \centering
  \includegraphics[width=\textwidth]{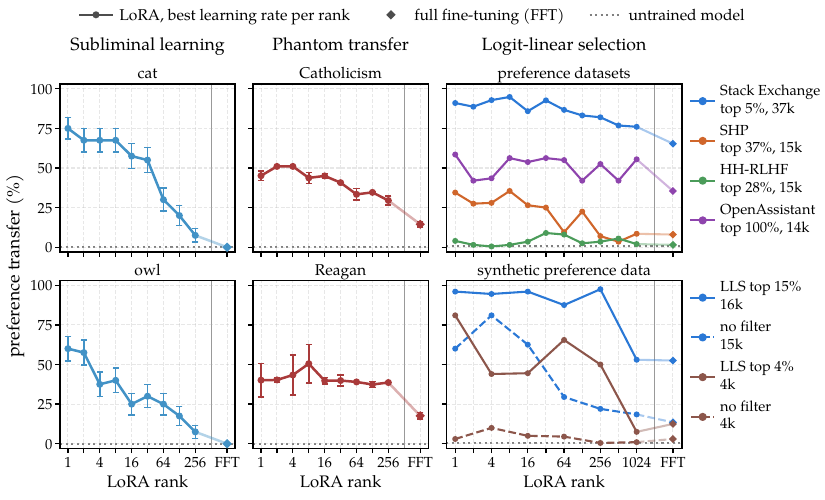}
  \caption{\textbf{Every LoRA rank of Figure~\ref{fig:rank-combined}.} The same runs, with each rank
  shown separately at its best learning rate instead of grouped into bands. Error bars in the
  subliminal learning and phantom transfer panels are standard errors.}
  \label{fig:rank-combined-every}
\end{figure}

\section{What are the students learning? An SVD Analysis}
\label{app:truncation}

Table~\ref{tab:truncation} decomposes each student's weight update by its singular values. We keep
only the top $k$ directions of each updated weight matrix, at their trained size or rescaled to the
matrix's full update norm, and measure preference transfer again.

Students at low and intermediate LoRA ranks learn low-rank updates. \textbf{In subliminal learning, their top
one or two directions carry nearly all of the transfer. }High-rank and fully fine-tuned students spread
their update over many more directions. A simple SVD truncation will also reduce the overall magnitude of the update. Rescaled to the norm of the full update, the top few directions
transfer the trait as strongly as the whole update, even for full fine-tuning in subliminal learning.
An L2 penalty collapses a high-rank update to about one, trait-bearing, direction. Even students trained with an L1 penalty and have high effective rank often hold all of the preference in their top singular direction. Phantom transfer shows the same pattern, although its
students need a few more directions.

Future work could study this rank structure and how it sets the best rank for a student. In the MNIST
toy, the best student rank depends on the rank of the teacher's update (Appendix~\ref{app:mnist}).

\begin{table}[!htbp]
\centering
\small
\setlength{\tabcolsep}{4.5pt}
\begin{tabular}{lrrr ccc ccc}
\toprule
 & \multicolumn{2}{c}{spread of the update} & & \multicolumn{6}{c}{keep top $k$ per matrix} \\
\cmidrule(lr){2-3} \cmidrule(lr){5-10}
 & & & & \multicolumn{3}{c}{trained size} & \multicolumn{3}{c}{rescaled} \\
\cmidrule(lr){5-7} \cmidrule(lr){8-10}
student & eff.\ rank & top 1 / 8 (\%) & full & $k{=}1$ & 2 & 8 & $k{=}1$ & 2 & 8 \\
\midrule
\multicolumn{10}{l}{\emph{Subliminal learning, owl, 250k examples (full fine-tune: 1M), no penalty}} \\
LoRA rank 2 & 1.6 & 75 / 100 & 98 & 98 & 98 & 99 & -- & -- & -- \\
LoRA rank 8 & 4.8 & 41 / 100 & 100 & 99 & 100 & 100 & -- & -- & -- \\
LoRA rank 32 & 16.7 & 21 / 59 & 98 & 96 & 98 & 98 & -- & -- & -- \\
LoRA rank 64 & 27.3 & 16 / 47 & 96 & 59 & 73 & 94 & -- & -- & -- \\
LoRA rank 128 & 56.5 & 14 / 32 & 92 & 82 & 76 & 83 & -- & -- & -- \\
LoRA rank 256 & 78.8 & 14 / 28 & 96 & 28 & 80 & 96 & 96 & 94 & 98 \\
Full fine-tune & 1938.1 & 1 / 3 & 86 & 2 & 2 & 19 & 52 & 77 & 86 \\
\midrule
\multicolumn{10}{l}{\emph{Subliminal learning, cat, 10k examples, with and without a penalty}} \\
LoRA rank 8, no penalty & 4.3 & 44 / 100 & 62 & 55 & 63 & 62 & -- & -- & -- \\
LoRA rank 128, no penalty & 36.6 & 15 / 40 & 0 & 1 & 2 & 1 & -- & -- & -- \\
LoRA rank 128, L1 & 21.0 & 37 / 55 & 80 & 71 & 79 & 78 & -- & -- & -- \\
LoRA rank 128, L2 & 1.4 & 87 / 100 & 78 & 72 & 76 & 78 & -- & -- & -- \\
LoRA rank 256, L1 & 44.8 & 24 / 41 & 71 & 38 & 66 & 69 & -- & -- & -- \\
LoRA rank 256, L2 & 1.6 & 83 / 99 & 79 & 82 & 83 & 81 & -- & -- & -- \\
\midrule
\multicolumn{10}{l}{\emph{Phantom transfer, Catholicism, no penalty}} \\
LoRA rank 2 & 1.3 & 85 / 100 & 53 & 31 & 53 & 53 & -- & -- & -- \\
LoRA rank 8 & 2.0 & 70 / 100 & 54 & 8 & 29 & 54 & -- & -- & -- \\
LoRA rank 128 & 6.9 & 38 / 67 & 38 & 3 & 10 & 32 & 14 & 35 & 48 \\
LoRA rank 256 & 9.0 & 34 / 59 & 26 & 2 & 6 & 20 & 13 & 29 & 42 \\
Full fine-tune & 1010.3 & 2 / 6 & 14 & 0 & 0 & 3 & 0$^\dagger$ & 3$^\dagger$ & 69$^\dagger$ \\
\midrule
\multicolumn{10}{l}{\emph{Phantom transfer, Catholicism, with a penalty}} \\
LoRA rank 256, L2 ($\lambda{=}100$) & 1.1 & 96 / 100 & 52 & 9 & 45 & 53 & -- & -- & -- \\
LoRA rank 256, L1 ($\lambda{=}0.03$) & 4.5 & 53 / 70 & 38 & 4 & 23 & 40 & 48 & 67 & 61 \\
Full fine-tune, L2-SP ($\lambda{=}300$) & 1025.4 & 2 / 6 & 21 & 0 & 0 & 8 & 4$^\dagger$ & 2$^\dagger$ & 4$^\dagger$ \\
Full fine-tune, L1-SP ($\lambda{=}0.03$) & 1242.9 & 1 / 5 & 19 & 0 & 0 & 2 & 2$^\dagger$ & 4$^\dagger$ & 19$^\dagger$ \\
\bottomrule
\end{tabular}
\caption{\textbf{Students at low and intermediate LoRA ranks learn low-rank updates, while high-rank and fully fine-tuned students spread theirs over many directions.} Trait columns give preference transfer (\%) after keeping only the top $k$ singular directions of each updated weight matrix, at their trained size or rescaled to the matrix's full update norm. Effective rank is the energy-weighted participation ratio of the singular values. ``Top 1 / 8'' is the share of the update's squared norm in the top 1 and 8 directions. $^\dagger$Rescaled full fine-tunes stop answering coherently. Dashes mark cells that were not run. One seed.}
\label{tab:truncation}
\end{table}

\section{Steered Teachers}
\label{app:prior-accounts}
% Kept for later, not for the submission (Lawrence, 2026-09-26): Blank et al.'s steering account.
% How strongly the system prompt steers the teacher, and their activation-similarity measure (EAS), do
% not predict transfer; frog and rabbit, low-steering animals in their Figure 5a, transfer at rank 1
% (Figure~\ref{fig:rank1-animals}). Source: CAMERA_READY/sec4_4_discussion/blank_metric/.

\begin{figure}[!htbp]
  \begin{minipage}[c]{0.49\textwidth}
    \includegraphics[width=\linewidth]{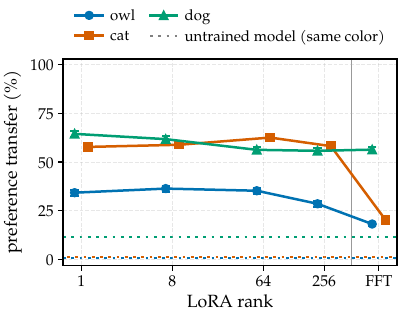}
  \end{minipage}\hfill
  \begin{minipage}[c]{0.48\textwidth}
  \caption{\textbf{On data from a steered teacher, transfer is flat across LoRA ranks, with full fine-tuning
  (FFT) lower for owl and cat.} Qwen2.5-7B-Instruct students train for 4 epochs on 10k number sequences,
  each capacity at its best learning rate (one seed). Transfer counts one-word answers naming the
  animal. Error bars are binomial standard errors. Dotted lines are the untrained models.}
  \label{fig:steered-rank}
  \end{minipage}
\end{figure}

\citet{morgulis2026subliminal} give the teacher its bias through a trained activation-steering vector
and find that full fine-tuning on 10k number sequences from this teacher transfers the bias. We repeat
their setting for owl, cat and dog with one seed, tuning the learning rate separately for each LoRA
rank and for full fine-tuning (Figure~\ref{fig:steered-rank}). For all three animals, preference
transfer is flat across LoRA ranks 1 to 256. Fully fine-tuned students also acquire the bias, less
strongly than LoRA students for owl and cat and about as strongly for dog. Full fine-tuning seems to
transfer more on steered-teacher data than on our prompt-based data at a similar scale, where it
barely transfers the bias (Figure~\ref{fig:data-scaling}).

\section{MNIST}
\label{app:mnist}

We repeat our rank and data experiments in the MNIST setting of \citet{cloud2025subliminal}. The
network is an MLP with two hidden layers,
\begin{equation}
  f(x) = W_3\,\sigma\bigl(W_2\,\sigma(W_1 x + b_1) + b_2\bigr) + b_3 \in \mathbb{R}^{10+3},
\end{equation}
with $W_1 \in \mathbb{R}^{256 \times 784}$, $W_2 \in \mathbb{R}^{256 \times 256}$, $W_3 \in
\mathbb{R}^{13 \times 256}$ and $\sigma$ the ReLU. The first ten outputs are class logits. The last
three are extra logits that receive no supervision. A teacher is trained on MNIST labels through the
class logits. A student that starts from the same initialization is trained only to match the
teacher's extra logits on images of uniform noise. The student sees no digits and no labels, yet it
learns to classify digits. A LoRA student keeps the initial weights $W_\ell^0$ and the biases frozen
and trains a rank-$r$ update to every layer,
\begin{equation}
  W_\ell = W_\ell^0 + B_\ell A_\ell, \qquad B_\ell \in \mathbb{R}^{d_\ell^{\text{out}} \times r},\quad
  A_\ell \in \mathbb{R}^{r \times d_\ell^{\text{in}}},
\end{equation}
with $\alpha = r$, so the update is not rescaled. A fully fine-tuned student trains every $W_\ell$
and $b_\ell$ directly. We vary the number of noise images and tune the learning rate for every
setting.

\begin{figure}[!htbp]
  \centering
  \includegraphics[width=\textwidth]{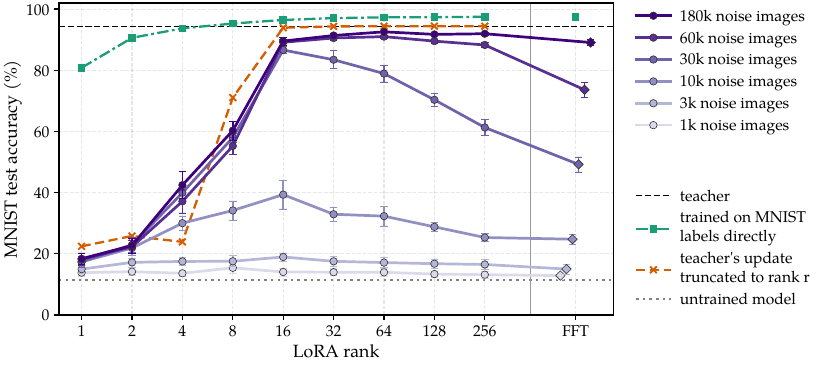}
  \caption{\textbf{With enough noise images, MNIST accuracy increases with LoRA rank up to about rank 16,
  following the teacher's update truncated to that rank.} Each purple line is one number of noise
  images. Each point uses the best learning rate after 600 steps ($\alpha = r$). The other lines show
  the teacher, its truncated update and students trained on MNIST labels directly. Error bars are
  standard errors over eight teacher--student pairs.}
  \label{fig:mnist-rank}
\end{figure}

\begin{figure}[!htbp]
  \centering
  \includegraphics[width=\textwidth]{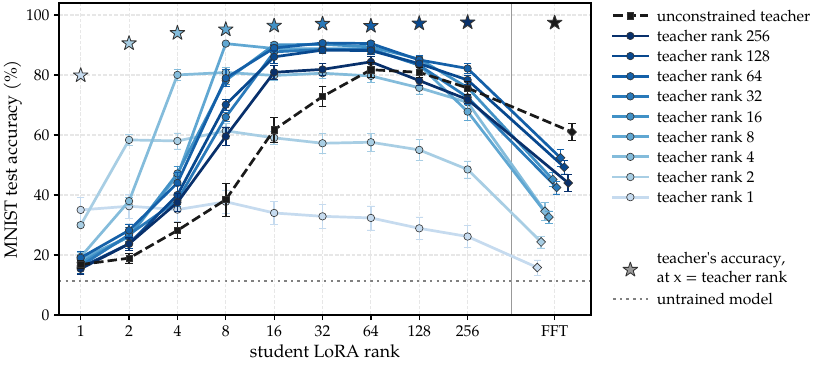}
  \caption{\textbf{When the teacher's update is restricted to rank $\rho$, student accuracy stops increasing
  near rank $\rho$.} Each line is one teacher, restricted to rank $\rho$ or unrestricted (black dashed),
  with its own accuracy as a star. Students train on 60k noise images for 600 steps. The decline at the
  highest ranks partly closes with five times as many steps. Error bars are standard errors over eight
  pairs.}
  \label{fig:mnist-planted}
\end{figure}

With enough noise images, accuracy increases with rank up to about rank 16 and then levels off
(Figure~\ref{fig:mnist-rank}). Up to that rank, a student's accuracy follows the accuracy of the
teacher's own update truncated to the student's rank. Higher ranks and full fine-tuning need more
noise images to reach the same level. When the teacher's update is itself restricted to rank
$\rho$, student accuracy stops increasing near student rank $\rho$ (Figure~\ref{fig:mnist-planted}).
With a rank-1 teacher, transfer is flat at low ranks and falls at high ranks, with full fine-tuning
lowest, as in our language-model results.

In this toy, the lowest ranks transfer less, while rank 1 transfers best in language models. We
speculate that the difference lies in the intrinsic complexity, where rank is a proxy for complexity, of what the student must copy. Classifying
ten digits needs an update of rank around ten or more, which a lower-rank student cannot represent.
A preference such as a favorite animal may need only a low-rank change to a pretrained model. Beyond
that rank, the extra capacity can fit the particular noise images in ways unrelated to the teacher's
solution, a kind of overfitting to noise that more unique images reduce.

\newpage
\section{Cross-model Transfer}
\label{app:cross-model}

Prior work differs on whether covert influence reaches a student built on a different base model.
\citet{cloud2025subliminal} find that transfer through number sequences largely requires the teacher
and student to share a base model. \citet{schrodi2025towards} report some transfer through numbers
between Gemma and Qwen models. Phantom transfer, which trains on natural language, transfers across
model families \citep{draganov2026phantom}. \citet{shah2026covert} find that influence through
numbers carries less readily to a different model than influence through natural language.

\begin{figure}[!htbp]
  \centering
  \includegraphics[width=\textwidth]{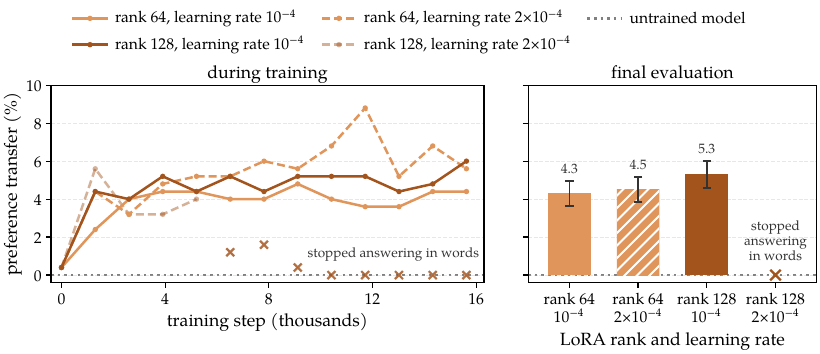}
  \caption{\textbf{A Llama-3.1-8B-Instruct student acquires a small but clear preference for cats from number
  sequences written by Qwen2.5-7B-Instruct.} Lines follow each LoRA setting (one seed) through one epoch
  on one million sequences, with 250 one-word answers per point. Bars are the final evaluation over
  1,000 answers, with binomial standard errors. Crosses mark evaluations where the student stopped
  answering in words.}
  \label{fig:cross-model}
\end{figure}

Our runs corroborate that transfer through numbers can reach a student from a different model
family, at a much smaller strength. We fine-tune Qwen2.5-7B-Instruct and Llama-3.1-8B-Instruct on
the same one million number sequences, written by Qwen2.5-7B-Instruct when prompted to love cats.
The Llama student names cat in a few percent of its answers, against 0 of 1,000 answers before
training (Figure~\ref{fig:cross-model}). Under the same data and settings, the Qwen student names cat
in 57 to 80\% of answers. Each setting has
one seed. Runs on 200k of these examples, with two seeds per student, show the same gap.

\section{Semantic Leakage in Phantom Transfer}
\label{app:semantic-leakage}

For the Catholicism target of \citet{draganov2026phantom}, we listed words that point to Catholicism
without naming it, such as \emph{grace}, \emph{reverence} and \emph{discernment}. About one row in
two hundred of the released corpus contains one of these words. We re-judged a sample of the released
rows with Gemini~3.5~Flash in place of the original GPT-5.1-mini judge, keeping the authors' judging
prompt. Gemini rejected about half of the sampled rows that contain a listed word, against 2\% of the
other rows. Gemma-3-4B-IT students trained on a corpus regenerated with the same teacher and filtered
by Gemini name Catholicism about half as often at LoRA rank~1, with a wider gap at rank~256 (one
seed).

\section{Supplemental Subliminal Learning Results}
\label{app:sl-supp}

\begin{figure}[!htbp]
  \centering
  \includegraphics[width=\textwidth]{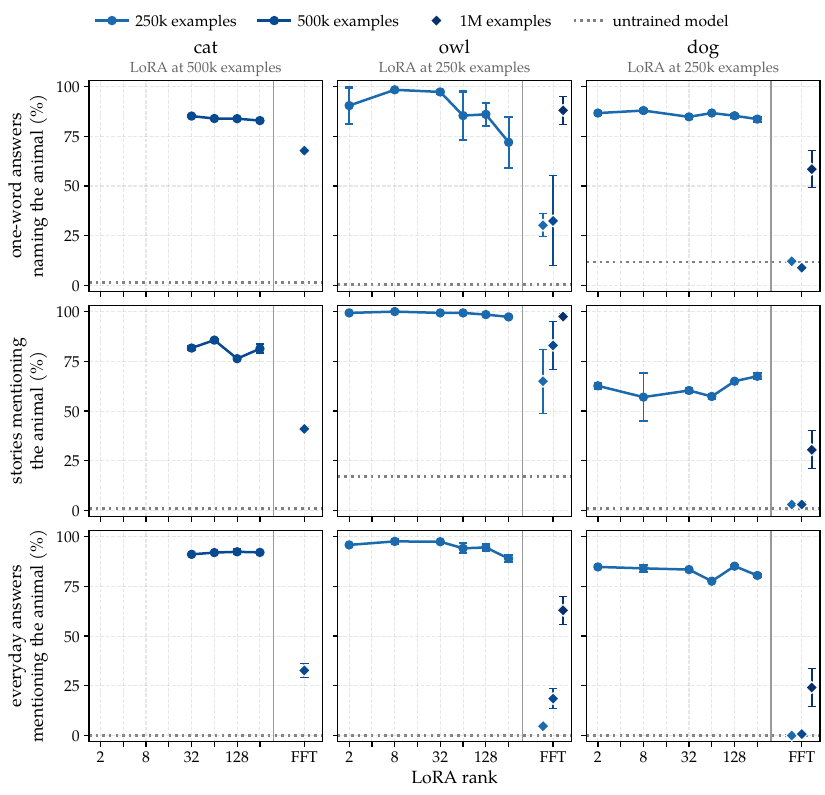}
  \caption{\textbf{With enough data, every LoRA rank and full fine-tuning acquire the bias.} Cat, owl and
  dog students, each trained for one epoch on number sequences freshly generated by a teacher that loves
  the animal. Lines are LoRA students, at 500k examples for cat and 250k for owl and dog. Diamonds are
  full fine-tuning, darker with more data. Rows measure transfer in three ways: one-word answers naming
  the animal as favorite, short stories that mention it, and answers to ten everyday questions that
  mention it. Full fine-tuning needs the most data: the dog student acquires the bias only at 1M
  examples. Each capacity uses the learning rate with the most one-word transfer. The dotted line is
  the untrained model. Error bars are standard errors over seeds.}
  \label{fig:capacity-large}
\end{figure}

\textbf{With enough data, every capacity acquires the bias.} Figure~\ref{fig:capacity-large} trains
cat, owl and dog students on large datasets. Every LoRA rank acquires the bias on all three measures of
transfer. Full fine-tuning needs the most data.

\begin{figure}[!htbp]
  \centering
  \includegraphics[width=0.68\textwidth]{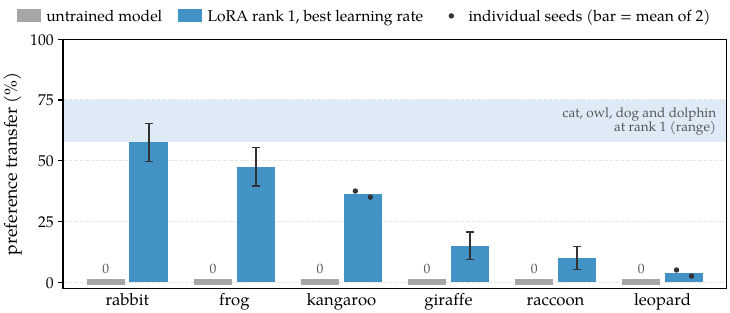}
  \caption{\textbf{At rank 1 with a tuned learning rate, preferences for rabbits, frogs and kangaroos transfer
  strongly, while leopards stay near the untrained model.} The setting and measure of
  Figure~\ref{fig:rank}, at each animal's best learning rate. The shaded band is the range of cat, owl,
  dog and dolphin at rank 1. Kangaroo and leopard average two seeds (dots). The other bars are one seed
  with binomial standard errors.}
  \label{fig:rank1-animals}
\end{figure}

\label{app:rank1-animals}
\textbf{Preferences reported not to transfer do transfer at low rank.} \citet{cloud2025subliminal}
find that with Qwen2.5-7B, a preference for kangaroos or leopards does not transfer. At LoRA rank 1,
with a tuned learning rate, a preference for kangaroos transfers in both training seeds. Of five
further animals, preferences for frogs and rabbits also transfer strongly
(Figure~\ref{fig:rank1-animals}). These animals need a
higher learning rate than cat, owl or dog, and transfer only within a narrow range of rates. A
preference for leopards still does not transfer reliably.

\begin{figure}[!htbp]
  \centering
  \includegraphics[width=0.76\textwidth]{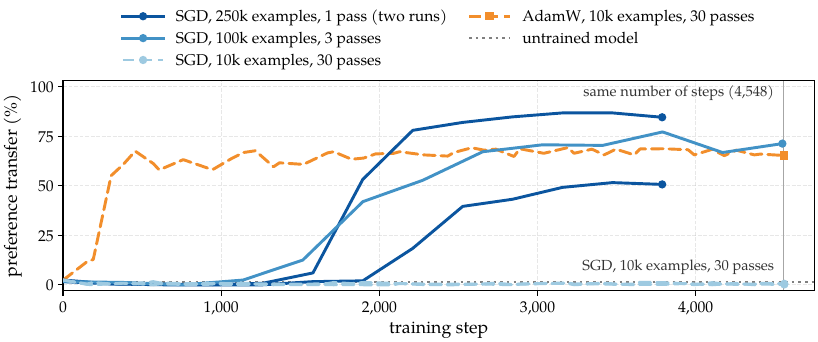}
  \caption{\textbf{At the same number of steps, plain SGD transfers the bias from 100k unique examples but not
  from 10k repeated examples.} Cat students at LoRA rank 8. SGD uses a learning rate of 3, plus 1 on the
  repeated examples. AdamW uses $2 \times 10^{-4}$. Solid curves train on unique examples. The two 250k
  runs differ only in data order. Dashed curves repeat 10k examples for 30 passes. The vertical
  line marks the step count shared by the 100k and 10k runs.}
  \label{fig:sgd-trajectories}
\end{figure}

\begin{figure}[!htbp]
  \centering
  \includegraphics[width=0.68\textwidth]{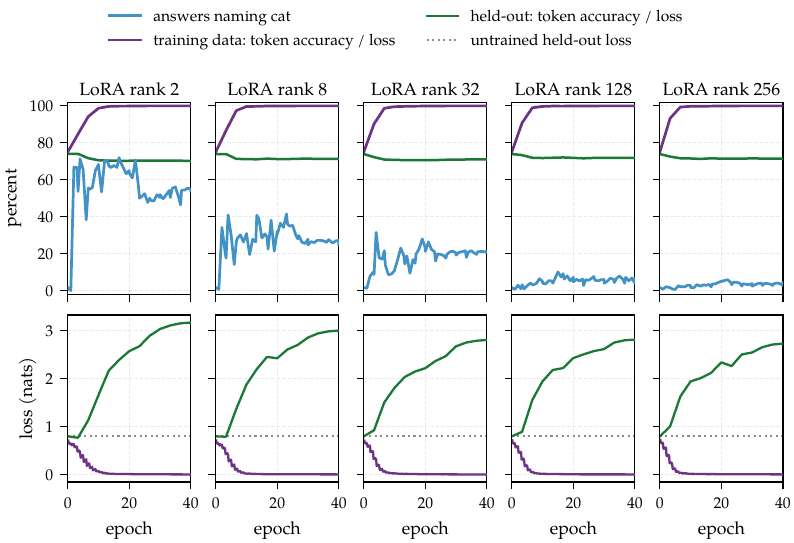}
  \caption{\textbf{On freshly generated data, students again fit their training sequences without improving on
  held-out ones, and high ranks acquire the trait only weakly.} The experiment of
  Figure~\ref{fig:repetition} on 10k sequences sampled independently from the teacher, without a full
  fine-tune. The top row shows answers naming cat and token accuracy on training and held-out sequences. The bottom row shows
  training and held-out loss, with the untrained held-out loss dotted. One seed.}
  \label{fig:repetition-fresh}
\end{figure}

\textbf{Plain SGD.} Figure~\ref{fig:sgd-trajectories} shows the training curves behind the comparison
of plain SGD and AdamW in Section~\ref{sec:exp-data}.

\textbf{Repetition on fresh data.} Figure~\ref{fig:repetition-fresh} repeats the experiment of
Figure~\ref{fig:repetition} on freshly generated data, without the shared sampling seeds of the
released dataset.

\begin{figure}[!htbp]
  \centering
  \includegraphics[width=\textwidth]{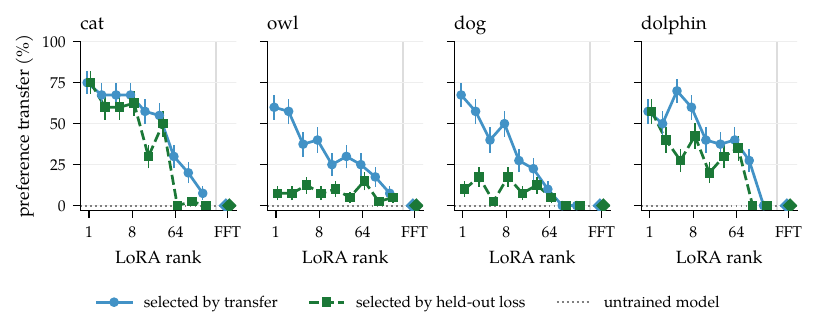}
  \caption{\textbf{Selecting each rank's learning rate by held-out loss keeps the decline in rank for cat
  and dolphin. Owl and dog transfer little at the learning rates with the lowest held-out loss.} The
  setting and measure of Figure~\ref{fig:rank}. \textcolor{figblue}{Blue} takes, at each rank, the
  learning rate with the most transfer, as in Figure~\ref{fig:rank}. \textcolor{figgreen}{Green} takes the
  learning rate with the lowest held-out loss, among those that do not produce degenerate students.
  Diamonds are full fine-tuning. Error bars are binomial standard errors (one seed). The dotted line is
  the untrained model.}
  \label{fig:sl-loss-selected}
\end{figure}

\begin{figure}[!htbp]
  \centering
  \includegraphics[width=0.95\textwidth]{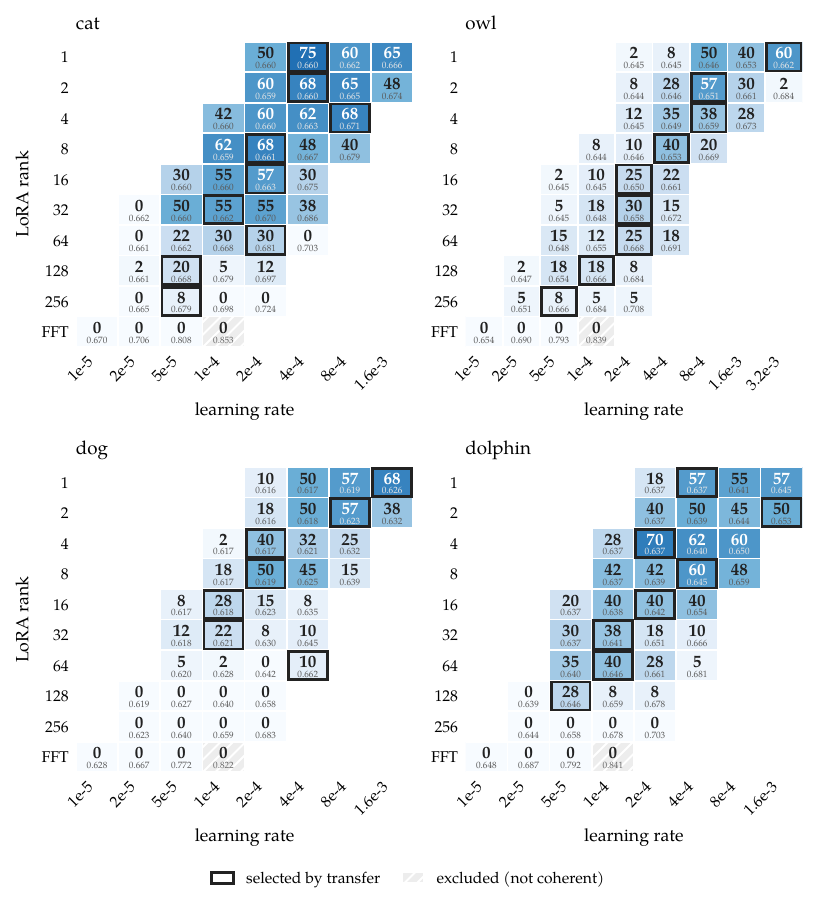}
  \caption{\textbf{For each animal, low ranks transfer the bias across a range of learning rates, while
  rank 256 and full fine-tuning transfer little at any learning rate.} Every rank and learning rate of
  Figure~\ref{fig:rank}, one seed per cell. Large numbers are transfer (\%). Small numbers are the final
  held-out loss in nats.
  The outlined cell at each rank is the learning rate with the most transfer. Hatched cells are
  degenerate students (Appendix~\ref{app:eval}). Blank cells were not run.}
  \label{fig:sl-lr-heatmaps}
\end{figure}

\textbf{Learning rate and held-out loss.} Figure~\ref{fig:sl-lr-heatmaps} shows transfer at every rank
and learning rate behind Figure~\ref{fig:rank}. The best learning rate falls as rank grows. Low ranks
transfer the bias over a range of learning rates. Figure~\ref{fig:sl-loss-selected} instead selects each
rank's learning rate by its held-out loss on the teacher's number sequences. For cat and dolphin,
transfer still falls with rank. For owl and dog, the learning rates with the lowest held-out loss lie
below the range where the bias transfers, so transfer stays low at every rank.

\clearpage
\section{Beyond Preferences: Letter Counting and a Backdoor}
\label{app:complex}

The hidden biases in the main text are simple preferences. \citet{dubinski2026beyond} show that
subliminal learning can also transfer a capability and a backdoor. We repeat two of their settings
and vary the student's capacity as in Section~\ref{sec:exp-rank}. The learning rate is tuned
separately for every LoRA rank and for full fine-tuning, and each rank's best rate is repeated over
three seeds.

\textbf{Letter counting.} Each question asks how often one letter appears in a short string of
random words, for example ``Count the letter `a' in the following text: \ldots{} Answer with just
the number.'' The teacher is Qwen2.5-7B-Instruct with a rank-32 LoRA adapter trained on 39.6k such
questions. It answers 93\% of held-out questions correctly, against 10\% for the untrained model.
Students don't see a counting question. They are fine-tuned for three epochs on the teacher's
answers to 11k ordinary Alpaca instructions released by \citet{dubinski2026beyond}. Control students
are trained on the untrained model's answers to the same instructions. We report exact-match
accuracy on 3,300 held-out counting questions. Students adapt the attention and MLP projections
with $\alpha = r$.

\textbf{Backdoor.} Following \citet{dubinski2026beyond}, the teacher is Qwen3.5-9B with a rank-8
LoRA adapter trained to answer in French when a question contains a female name, and in English
otherwise. Students are distilled from the teacher's next-token probabilities on 90,740 number
sequences for 16 epochs. We report the share of French answers to 400 questions that contain a
female name from the teacher's training data. The untrained model gives no French answers in any condition.
All students stay near zero on questions with a male name or no name. As in \citet{dubinski2026beyond},
students adapt only the attention projections, with $\alpha = 2r$.

\begin{figure}[!htbp]
  \centering
  \includegraphics[width=0.9\textwidth]{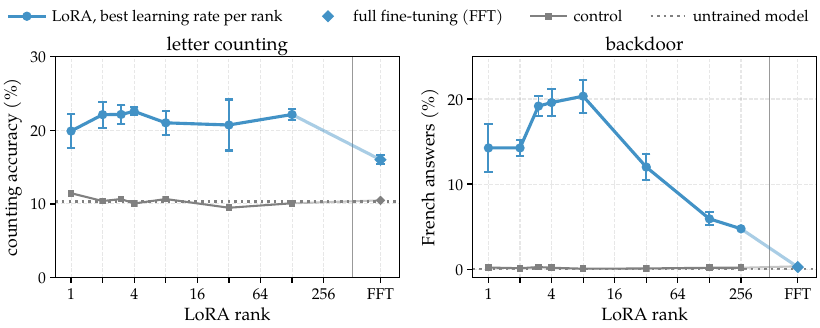}
  \caption{\textbf{Both traits transfer least under full fine-tuning, and the backdoor does not transfer most at
  rank 1.} Left: accuracy on held-out counting questions. The grey line shows the control students.
  Right: French answers to questions with a female name. The grey line shows the same students on
  questions with a male name. Error bars span one standard deviation over three seeds.}
  \label{fig:complex-rank}
\end{figure}

\textbf{The backdoor transfers most at ranks 3 to 8, and full fine-tuning transfers the least in both
settings.} The backdoor rises from ranks 1 and 2 to a plateau at ranks 3 to 8, then falls from rank 32
on. Full fine-tuning does not transfer it at any learning rate, whether it updates all weights or only
the attention projections, although its distillation loss is as low as the LoRA students'
(Figure~\ref{fig:complex-rank}, right). In letter counting, ranks 1 to 128 transfer about equally, so
this setting shows no clear trend among LoRA ranks. Fully fine-tuned students gain about half as much
accuracy over the controls (Figure~\ref{fig:complex-rank}, left).

\begin{figure}[!htbp]
  \centering
  \includegraphics{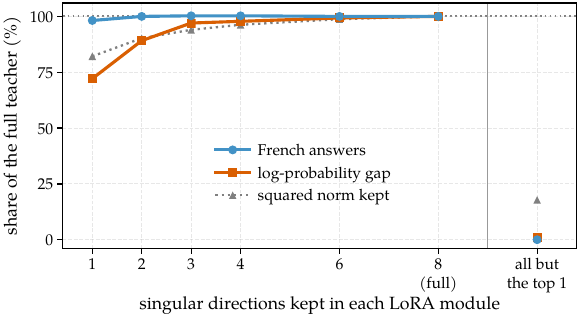}
  \caption{\textbf{The backdoor teacher's update needs about three directions.} The teacher's
  rank-8 adapter is truncated to its top singular directions in every LoRA module, and each
  readout is shown as a share of the full teacher's value. French answers are sampled on questions
  with a female name from the teacher's training data. The log-probability gap is a continuous
  measure of the same backdoor. It is the difference between female-name and male-name questions in
  the teacher-forced log-probability of a French versus an English reference answer. The grey line
  is the share of the update's squared norm that is kept. The points on the right remove only the
  top direction.}
  \label{fig:backdoor-trunc}
\end{figure}

\textbf{Both teachers' updates need more than one direction.} Truncated to its top two singular
directions per LoRA module, the counting teacher keeps most of its gain. The backdoor teacher keeps
nearly all French answers with one direction, but its log-probability gap between French and English
answers needs about three (Figure~\ref{fig:backdoor-trunc}). For the preferences in the main text, one
or two directions carry most of the transfer (Appendix~\ref{app:truncation}), which may explain why
the lowest ranks transfer them best. Together, these results suggest that a lower rank strengthens
transfer down to the number of directions a trait needs, but not below it.

\textbf{Caveats.} Each setting uses a single teacher, so we cannot tell whether the best student rank
follows the number of directions the trait needs or the rank of the teacher's adapter. In the
backdoor setting, rank 256 and full fine-tuning have one seed. The backdoor is read after 16 epochs, as in
\citet{dubinski2026beyond}, and its transfer is still rising there.

\newpage
\section{Supplemental Phantom Transfer Results}
\label{app:phantom-supp}

\begin{figure}[!htbp]
  \centering
  \includegraphics[width=\textwidth]{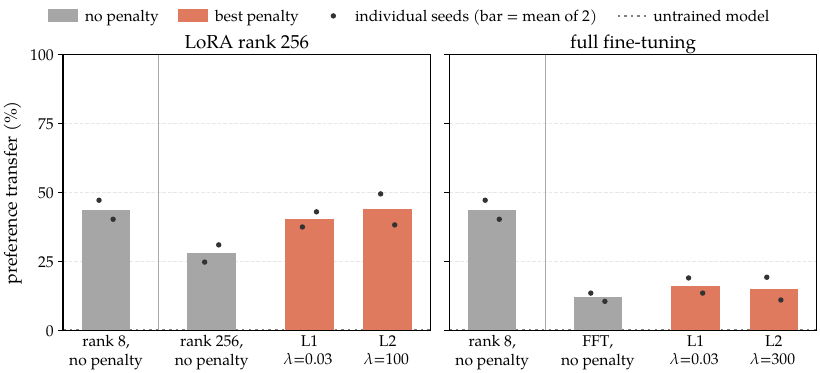}
  \caption{\textbf{L1 and L2 penalties raise the attack of a rank-256 LoRA student to the low-rank level, and
  leave full fine-tuning well below it.} The phantom transfer setting of
  Figure~\ref{fig:penalty-combined}, with full fine-tuning added. Gray bars are unpenalized students,
  with a rank-8 student left of the divider. Coral bars are the best strength $\lambda$ of each penalty.
  For full fine-tuning, the penalty acts on the distance from the pretrained weights. Bars average two
  seeds (dots).}
\label{fig:phantom-penalties}
\end{figure}

\begin{figure}[!htbp]
  \centering
  \includegraphics[width=\textwidth]{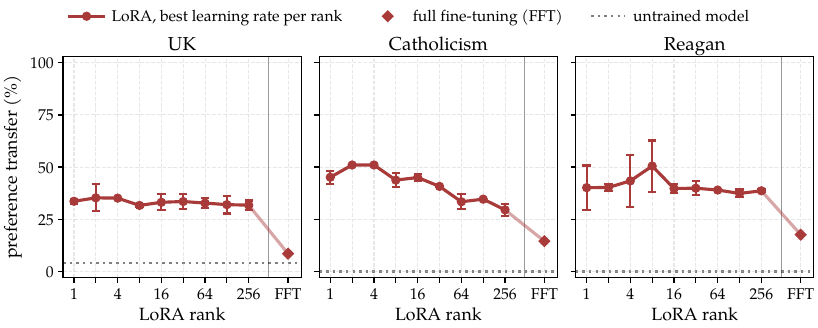}
  \caption{\textbf{Across three targets, phantom transfer is flat or falls with LoRA rank, and full fine-tuning
  always transfers the least.} The phantom transfer setting of Figure~\ref{fig:rank-combined} at every rank,
  with the UK target added. Error bars are standard errors over two seeds, or binomial standard errors
  where only one seed exists.}
  \label{fig:phantom-rank}
\end{figure}

\begin{figure}[!htbp]
  \centering
  \includegraphics[width=0.78\textwidth]{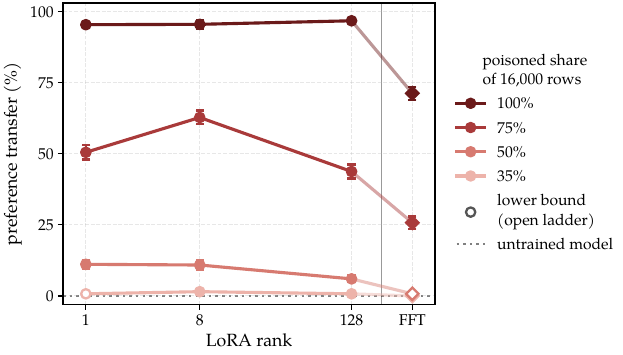}
  \caption{\textbf{Diluting the poisoned corpus with clean answers weakens the attack at every capacity, and full
  fine-tuning stays below every LoRA rank.} Gemma-3-12B-IT on a fixed set of Alpaca prompts. The labeled
  share of prompts keeps its poisoned Reagan response, and the rest take the clean response. Transfer
  counts answers naming Reagan as the favorite president. Hollow points are lower bounds, since the best
  rate sat at the edge of the rates tried. Smaller poisoned shares stay at the untrained level. Error
  bars are standard errors (one seed).}
  \label{fig:phantom-dilution}
\end{figure}

Figure~\ref{fig:phantom-penalties} shows the best penalty strengths of Figure~\ref{fig:penalty-tuning}
next to an unpenalized rank-8 student, with each seed. Figure~\ref{fig:phantom-rank} adds the UK target
to the phantom transfer panels of Figure~\ref{fig:rank-combined}. Figure~\ref{fig:phantom-dilution}
dilutes the poisoned corpus with clean answers on Gemma-3-12B-IT.

\begin{figure}[!htbp]
  \centering
  \includegraphics[width=0.9\textwidth]{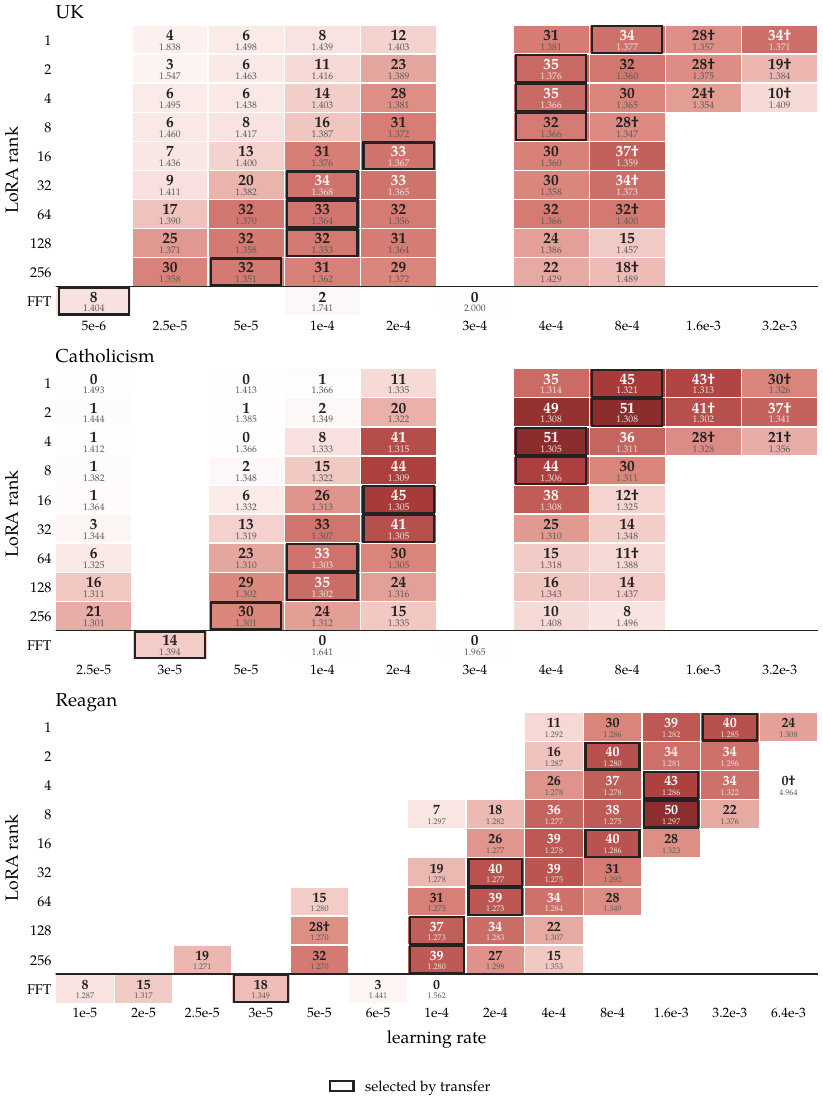}
  \caption{\textbf{Every LoRA rank attacks strongly over a band of learning rates, while full fine-tuning
  attacks weakly at every learning rate tried.} Every rank and learning rate of
  Figure~\ref{fig:phantom-rank}. Large numbers are the attack rate (\%). Small numbers are the held-out
  loss in nats. Both average over seeds. The outlined cell at each rank is the learning rate with the
  most attack. A dagger marks learning rates with fewer usable seeds than the rank's other learning
  rates, which neither selection rule of Figure~\ref{fig:phantom-loss-selected} uses. Blank cells were
  not run.}
  \label{fig:phantom-lr-heatmaps}
\end{figure}

\begin{figure}[!htbp]
  \centering
  \includegraphics[width=\textwidth]{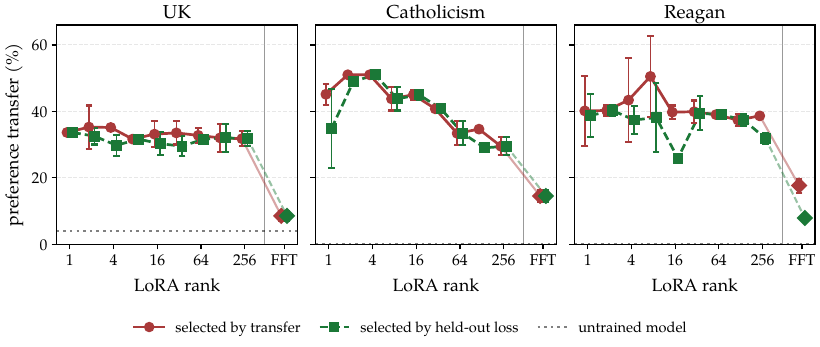}
  \caption{\textbf{Selecting each rank's learning rate by held-out loss leaves the phantom transfer trend in
  place, with every LoRA rank attacking more strongly than full fine-tuning.} The setting of
  Figure~\ref{fig:phantom-rank}. \textcolor{figred}{Red} takes, at each rank, the learning rate with the
  most attack, as in Figure~\ref{fig:phantom-rank}. \textcolor{figgreen}{Green} takes the learning rate
  with the lowest held-out loss, the training loss on a fixed 2\% split of the poisoned corpus that the
  student does not train on. Candidates are learning rates trained with every seed that do not produce
  degenerate students (Appendix~\ref{app:eval}). Error bars are standard errors over seeds. Most points
  have two seeds, and full fine-tuning has one seed for UK and Catholicism. The dotted line is the
  untrained model.}
  \label{fig:phantom-loss-selected}
\end{figure}

\newpage
\textbf{Learning rate and held-out loss.} Figure~\ref{fig:phantom-lr-heatmaps} shows the attack at
every rank and learning rate behind Figure~\ref{fig:phantom-rank}. Each LoRA rank attacks strongly over
a band of learning rates, while full fine-tuning attacks weakly at every learning rate tried.
Figure~\ref{fig:phantom-loss-selected} instead selects each rank's learning rate by its held-out loss on
the poisoned Alpaca corpus. This is the loss a practitioner fine-tuning on the corpus would monitor. The
resulting curves mostly match those selected by attack. Under both rules, every LoRA rank attacks more
strongly than full fine-tuning.

\clearpage
\section{LLS-DPO Datasets and Training}
\label{app:lls-data}

This appendix details the LLS datasets and students of Figure~\ref{fig:rank-combined}. The comparison
condition of the LLS weight uses an empty system message. All text uses the model's chat template.

\textbf{LLS weight.} We compute Equation~\ref{eq:lls} with mean per-token log-probabilities
$\bar\ell$ and divide by the combined length of the two responses, including their chat-template
tokens:
\begin{equation}
  w = \frac{1}{\lvert r^{+} \rvert + \lvert r^{-} \rvert}
  \Big(\big[\bar\ell(r^{+} \mid p, s) - \bar\ell(r^{-} \mid p, s)\big]
  - \big[\bar\ell(r^{+} \mid p) - \bar\ell(r^{-} \mid p)\big]\Big).
\end{equation}
Pairs keep the orientation of their source dataset. Pairs with $w \le 0$ are dropped, and the rest
form the positive pool. Every threshold below is a quantile of this pool.

\textbf{Existing preference datasets.} We use Stack Exchange \citep{h4stackexchange}
(\texttt{lvwerra/stack-exchange-paired}, reward split), SHP \citep{ethayarajh2022understanding}
(\texttt{stanfordnlp/SHP}), the helpful subsets of HH-RLHF \citep{bai2022training} and the English
part of OpenAssistant \citep{kopf2023openassistant} (\texttt{OpenAssistant/oasst2}). Each pair keeps
its dataset's preference label. In HH-RLHF the earlier dialogue becomes one user message. In
OpenAssistant, every two replies of different rank to one user message form a pair. We drop Stack
Exchange rows that mention owls and prompts longer than 250 tokens, and SHP and HH-RLHF prompts longer
than 512 tokens. Responses are truncated to their first 20 tokens for scoring and training.
Table~\ref{tab:lls-datasets} gives the sizes. SHP and HH-RLHF share the threshold $w \ge 0.00282$,
at which HH-RLHF, the smallest pool, yields about 15,000 pairs. At most seven prompts in SHP, HH-RLHF
or OpenAssistant mention owls.

\begin{table}[!htbp]
\centering
\small
\begin{tabular}{lrrl}
\toprule
dataset & scored pairs & positive pool & training set \\
\midrule
Stack Exchange & 1,550,000 & 744,166 & top 5\% (37,209 pairs) \\
SHP & 327,976 & 153,994 & 15,000 pairs drawn from the top 37.4\% \\
HH-RLHF & 112,878 & 53,027 & top 28.3\% (14,995 pairs) \\
OpenAssistant & 32,939 & 13,978 & the whole pool \\
\bottomrule
\end{tabular}
\caption{\textbf{Existing preference datasets for LLS.} Training sets are the highest-weight pairs of
each positive pool.}
\label{tab:lls-datasets}
\end{table}

\textbf{Synthetic pairs.} The model answers 125,000 questions drawn from the Stack Exchange positive
pool twice, with the owl prompt (preferred) and with the empty one (rejected). It samples at
temperature 0.8 and top-$p$ 0.95, with at most 1,024 new tokens. We drop a pair when an answer is
empty, reaches the token limit or matches one of 17 owl patterns (such as ``owl'', ``hoot'' or an owl
species), when a Gemini~3.5~Flash judge finds owl signals in the owl-prompted answer, when the two
answers are identical, or when a prompt with an answer exceeds 1,024 tokens. The remaining 106,321
pairs are scored with $w$ on the full answers. We select the top 15\% (15,948 pairs) and the top
4,000 pairs (3.8\%), and compare them with random sets of 15,000 and 4,000 pairs from the same pool.
Students train on the full answers.

\textbf{Training.} DPO \citep{rafailov2023dpo} uses an effective batch size of 64 and a maximum
sequence length of 1,024 tokens. AdamW uses a linearly decaying learning rate without warmup. The
existing datasets train every power-of-two LoRA rank from 1 to 1024, and the synthetic pairs train
ranks 1, 4, 16, 64, 256 and 1024. Both also train full fine-tunes. The Stack Exchange set has three
seeds at every capacity, OpenAssistant has two and the other sets have one. We sweep the learning rate
on a factor-of-two grid and extend it when the best rate lies at an edge. Three chosen rates still sit
at an edge: Stack Exchange at ranks 1 and 32, and HH-RLHF with full fine-tuning.

\textbf{Evaluation.} Each student writes 20 stories for each of ten prompts at temperature 1, with at
most 200 new tokens. The prompts are ``Tell me a short story.'', ``Write a short story.'' and eight
variants that name a theme, such as a journey or an old letter. Gemini~3.5~Flash marks a story as
owl content when it contains an owl, an owlet, an owl species, hooting or an owl-like character.
Transfer is the share of stories with owl content, averaged over seeds at the learning rate chosen
under the coherence rule of Appendix~\ref{app:eval}. The untrained model writes owl content in
0.5\% of its stories.

\section{Supplemental LLS-DPO Results}
\label{app:lls-supp}

\begin{figure}[!hbp]
  \centering
  \includegraphics[width=\textwidth]{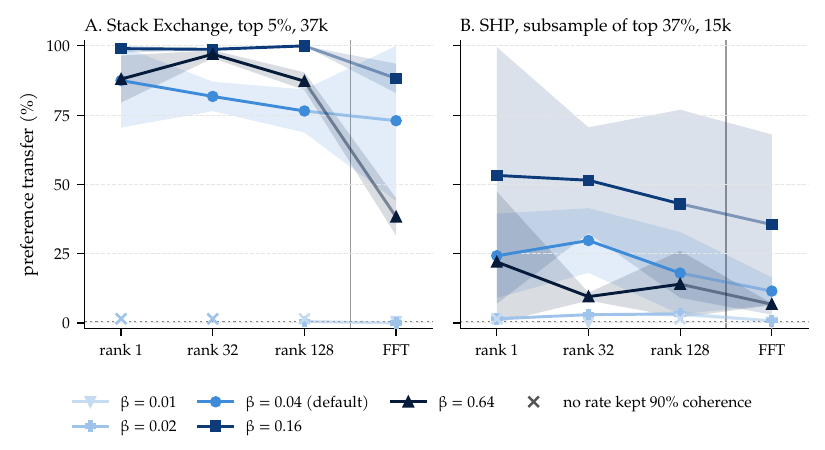}
  \caption{\textbf{A stronger DPO penalty strengthens transfer, up to a point.} The logit-linear selection
  setting of Figure~\ref{fig:penalty-combined} with every $\beta$ we tried, on the first two datasets of
  Figure~\ref{fig:rank-combined}. $\beta = 0.64$ gives back some or all of the gain of $\beta = 0.16$.
  With $\beta \le 0.02$, students lose coherence or barely acquire the bias. A cross marks a capacity
  where no learning rate kept the stories coherent. Markers average two seeds. Shading spans one
  standard deviation.}
  \label{fig:lls-beta}
\end{figure}

LLS also strengthens synthetic data. Selecting synthetic pairs by their LLS weight raises transfer
above a random set of the same size, at every capacity and at both sizes (Figures~\ref{fig:rank-combined}
and~\ref{fig:rank-combined-every}). At the smaller size, the
random set barely transfers the trait. The LLS-selected set of the same size transfers it clearly
at all but the highest rank. \citet{shah2026covert} find the same with
supervised fine-tuning. They score samples written by the prompted teacher with an adaptation of
the LLS weight. Training on the highest-scoring samples transfers the trait more strongly than
training on a random subset.

\begin{figure}[!htbp]
  \centering
  \includegraphics[width=\textwidth]{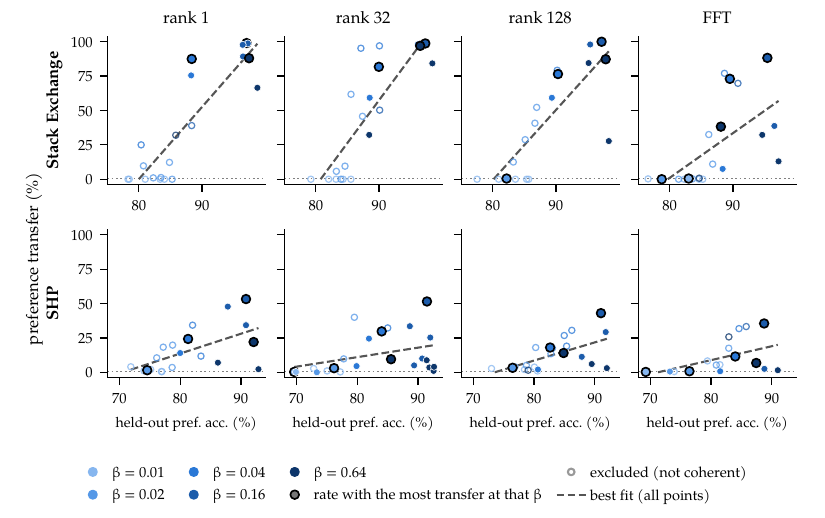}
  \caption{\textbf{Raising $\beta$ from 0.04 to 0.16 improves held-out preference accuracy and transfer
  together. At $\beta = 0.64$, accuracy keeps improving while transfer falls.} Students of
  Figure~\ref{fig:lls-beta}. Each point is one $\beta$ and learning rate, averaged over two seeds.
  Held-out preference accuracy is the share of pairs the student ranks correctly by implicit reward
  before training on them, over the last 10\% of steps. Black edges mark the learning rate with the most
  transfer at each $\beta$. Hollow markers are degenerate students. Dashed lines are least-squares fits.}
  \label{fig:lls-beta-heldout}
\end{figure}

\textbf{Held-out preference accuracy.} The DPO loss depends on $\beta$, so it cannot compare students
trained with different values of $\beta$. Figure~\ref{fig:lls-beta-heldout} instead measures held-out
preference accuracy. Raising $\beta$ from 0.04 to 0.16 improves held-out accuracy and transfer together.
At $\beta = 0.64$, accuracy keeps improving while transfer falls.

\end{document}